\documentclass[sigconf, nonacm]{acmart}
\usepackage{graphicx}
\usepackage{textcomp}
\usepackage{xcolor}
\usepackage{algorithm}
\usepackage[noend]{algpseudocode}
\usepackage{multirow}
\usepackage{colortbl}
\usepackage{booktabs}
\newcommand{\std}[1]{{\tiny$\pm$#1}}
\newcommand{\stdc}[1]{{\fontsize{4.6}{5}\selectfont$\pm$#1}}  % compact, for Tables 1 & 4
\definecolor{cGreen}{RGB}{200,230,201}
\definecolor{cYellow}{RGB}{255,249,196}
\definecolor{cOrange}{RGB}{255,224,178}
\definecolor{cRed}{RGB}{239,154,154}
\pdfoutput=1
\makeatletter
\newcommand{\linebreakand}{%
  \end{@IEEEauthorhalign}
  \hfill\mbox{ }\par\vspace{1em}
  \mbox{ }\hfill\begin{@IEEEauthorhalign}
}
\makeatother

\begin{document}

\title{SPARC: Accurate Photonic Matrix Multiplication via Representative Vectors}


\author{Yasuhiro Fujiwara}
\affiliation{%
  \institution{NTT, Inc.}
  \country{Japan}}
\email{yasuhiro.fujiwara@ntt.com}

\author{Tatsuhiro Nakamori}
\affiliation{%
  \institution{Keio University}
  \country{Japan}}
\email{tatsuhironm@keio.jp}

\author{Hideyuki Kawashima}
\affiliation{%
  \institution{Keio University}
  \country{Japan}}
\email{river@sfc.keio.ac.jp}

\begin{abstract}
Photonic neural network accelerators have emerged as a promising solution for
reducing the power consumption of deep neural network (DNN) inference
by executing matrix multiplication on analog photonic hardware.
However, their inference accuracy is limited by low-bit quantization
required for photonic processing and by analog photonic noise.
This paper proposes an approach for improving inference accuracy
by exploiting the high similarity among activation vectors in DNNs.
Specifically, our approach 
constructs a representative vector for each group of similar activation vectors,
computes its inner product digitally with high precision,
and offloads the residual inner products to the photonic accelerator.
Since the activation vectors within each group are highly similar,
the residual vectors have much smaller norms,
thereby reducing the effects of quantization errors and analog photonic noise.
Furthermore, our approach preserves the photonic inner-product computation pipeline,
allowing seamless integration into existing and future photonic accelerators
without hardware modification.
Experimental results on six representative DNN models demonstrate that
the proposed approach can improve inference accuracy across diverse network architectures and application domains.

\end{abstract}

\maketitle

\section{Introduction}
\label{sec:intro}

Deep neural networks (DNNs) have become a cornerstone of modern AI,
achieving remarkable success in applications such as computer vision,
natural language processing, and data mining
\cite{DBLP:journals/pieee/SzeCYE17}. 
The rapid growth of AI services has substantially increased power
consumption, particularly in data centers, 
where DNN inference accounts for a significant fraction of electricity consumption
\cite{DBLP:journals/suscom/DesislavovMH23}.
Moreover, the growing scale of modern neural networks has significantly
increased the computational cost of DNN inference
\cite{Xu2018ScalingFE}.
DNN inference is dominated by matrix multiplication
\cite{DBLP:journals/pieee/SzeCYE17}.
Let $\mathbf{A}$, $\mathbf{B}$, and $\mathbf{C}$ denote matrices of size
$N \times D$, $D \times M$, and $N \times M$, respectively.
The matrix multiplication is given by
$\mathbf{A}\mathbf{B}=\mathbf{C}$. 
Developing energy-efficient matrix multiplication approaches
has therefore become a key research direction \cite{DBLP:journals/access/MarinisCCA19,11284869}.

\begin{figure}[t]
\begin{center}
\small
\begin{tabular}{c}
\hspace{-1.5em} \includegraphics[width=85mm,clip]{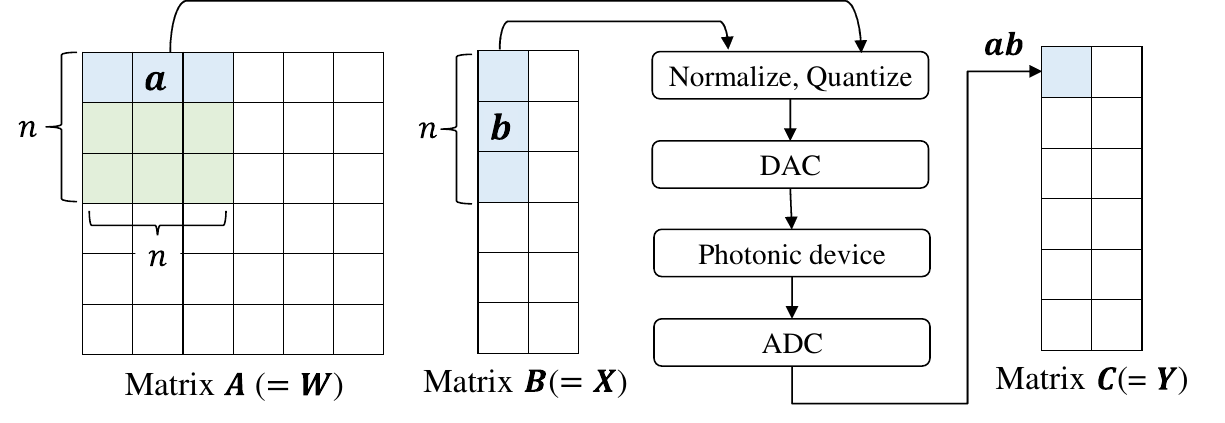}\vspace{0.0em}\\
\hspace{-1.5em} (1) Original photonic matrix-multiplication approach \vspace{1.0em}\\
\hspace{-1.5em} \includegraphics[width=85mm,clip]{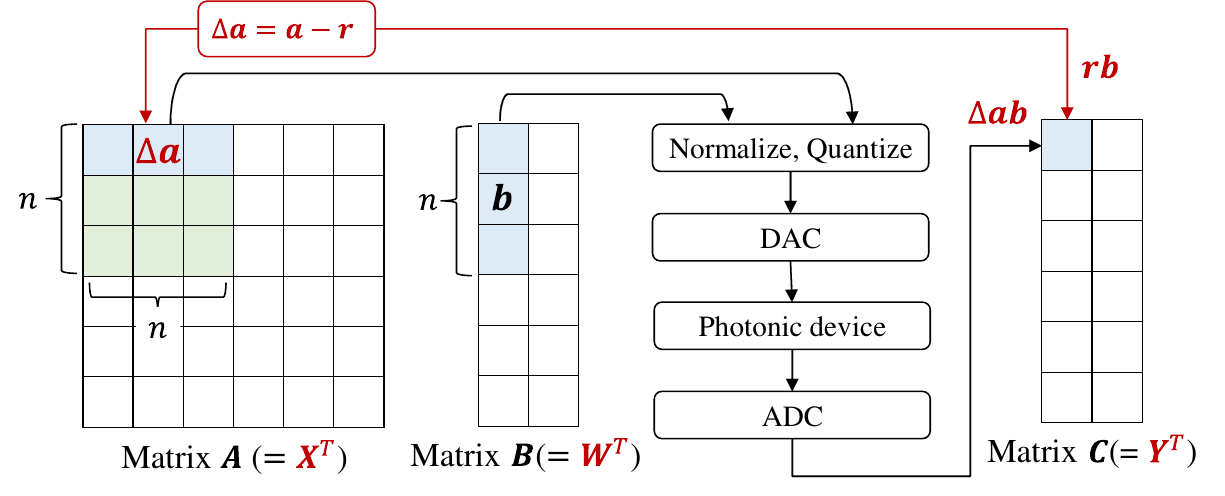}\vspace{0.0em}\\
\hspace{-1.5em} (2) Proposed photonic matrix-multiplication approach
\end{tabular}
\vspace{-1.2em}
\caption{Block-wise matrix multiplication.}
\label{fig:matrix_mult}
\normalsize
\end{center}
\vspace{-0.0em}
\end{figure}

Recently, Lightmatter introduced a universal photonic accelerator in Nature,
advancing photonic neural network accelerators from
laboratory-scale prototypes toward practical deployment 
for large-scale DNN inference \cite{RePEc:nat:nature:v:640:y:2025:i:8058:d:10.1038_s41586-025-08854-x}.
Rather than performing matrix multiplication entirely on power-hungry digital hardware, 
the accelerator offloads the computation to a low-power photonic analog device, 
thereby significantly reducing energy consumption \cite{Shen:2017fqu}.
Let $\mathbf{W}$ and $\mathbf{X}$ denote the weight matrix and activation
matrix of a DNN, respectively.
As illustrated in Figure~\ref{fig:matrix_mult}-1,
the accelerator computes
$\mathbf{WX}=\mathbf{Y}$ by mapping
$\mathbf{W}$, $\mathbf{X}$, and $\mathbf{Y}$
to $\mathbf{A}$, $\mathbf{B}$, and $\mathbf{C}$, respectively.
The matrix product $\mathbf{A}\mathbf{B}$ is decomposed
into vector inner products.
Specifically, matrix $\mathbf{A}$ is partitioned into $n \times n$ block matrices, and 
matrix $\mathbf{B}$ is partitioned into corresponding block column vectors of length $n$.
Each output element is computed as the inner product of
a row vector $\mathbf{a}$ from a block of $\mathbf{A}$
and the corresponding column vector $\mathbf{b}$ from $\mathbf{B}$.
Before photonic computation,
the vectors are normalized and quantized using
Adaptive Block Floating-Point (ABFP) 
\cite{RePEc:nat:nature:v:640:y:2025:i:8058:d:10.1038_s41586-025-08854-x,DBLP:journals/corr/abs-2205-06287},
converted into analog signals by Digital-to-Analog Converter (DAC),
processed on photonic analog hardware,
and finally digitized by Analog-to-Digital Converter (ADC).
By offloading the computationally intensive inner-product operations
to photonic hardware,
the accelerator achieves substantially higher energy efficiency than
conventional digital accelerators.

However, photonic computation inevitably suffers from numerical errors. 
Before photonic computation, floating-point values (e.g., $32$ bits)
must be quantized into low-bit representations (e.g., $6$ bits) for DAC,
introducing quantization errors, and
the limited precision of ADC (e.g., $8$ bits) further degrades inference accuracy
\cite{DBLP:journals/corr/abs-2205-06287}.
More importantly, analog photonic computation is inherently affected by
analog photonic noise \cite{RePEc:nat:nature:v:640:y:2025:i:8058:d:10.1038_s41586-025-08854-x}. 
Consequently, improving the accuracy of photonic matrix multiplication
has become a major challenge.

\begin{figure}
\begin{center}
\small
\begin{tabular}{cccc}
\includegraphics[height=18mm,clip]{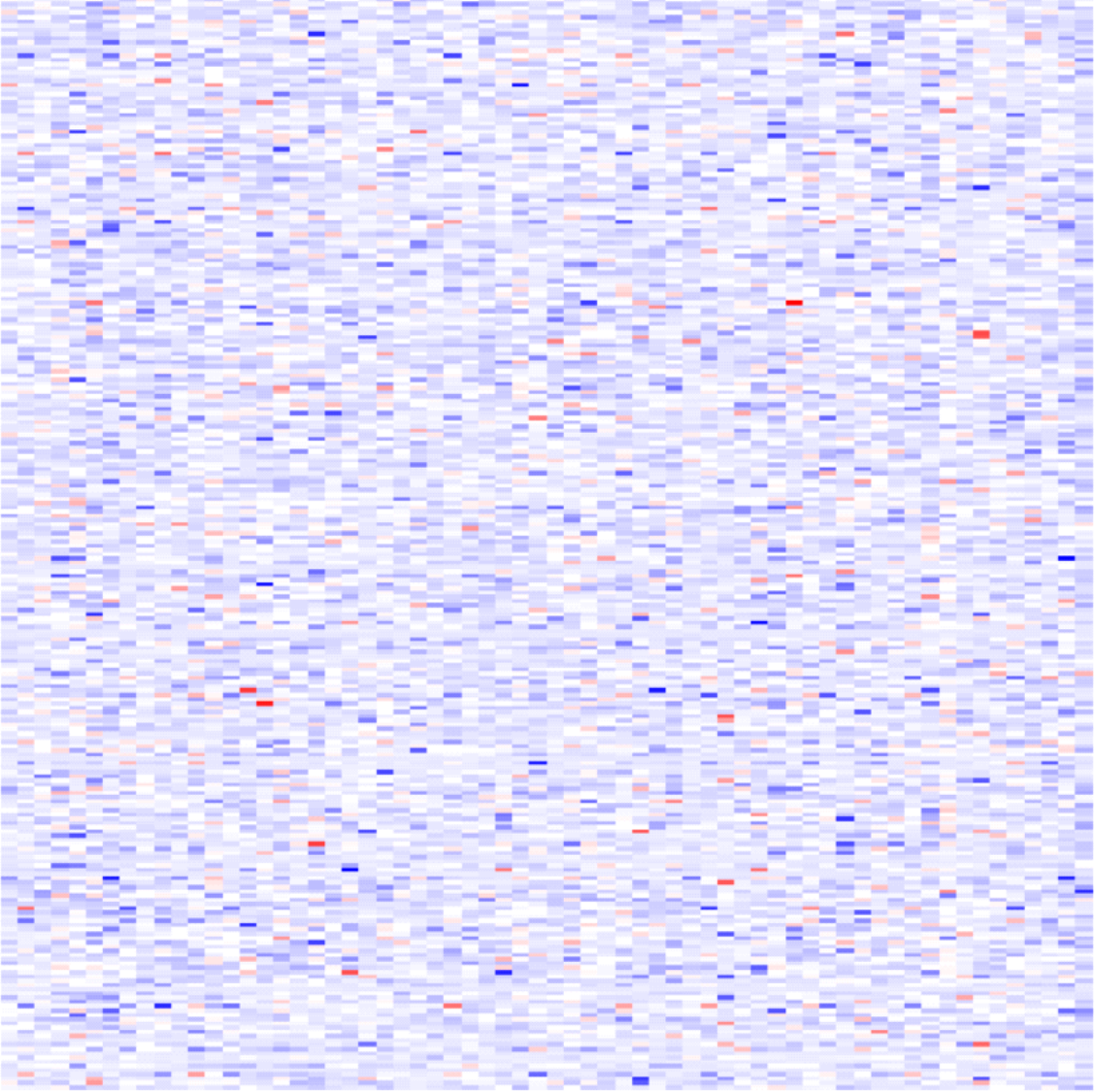}&
\hspace{-1.0em}
\includegraphics[height=18mm,clip]{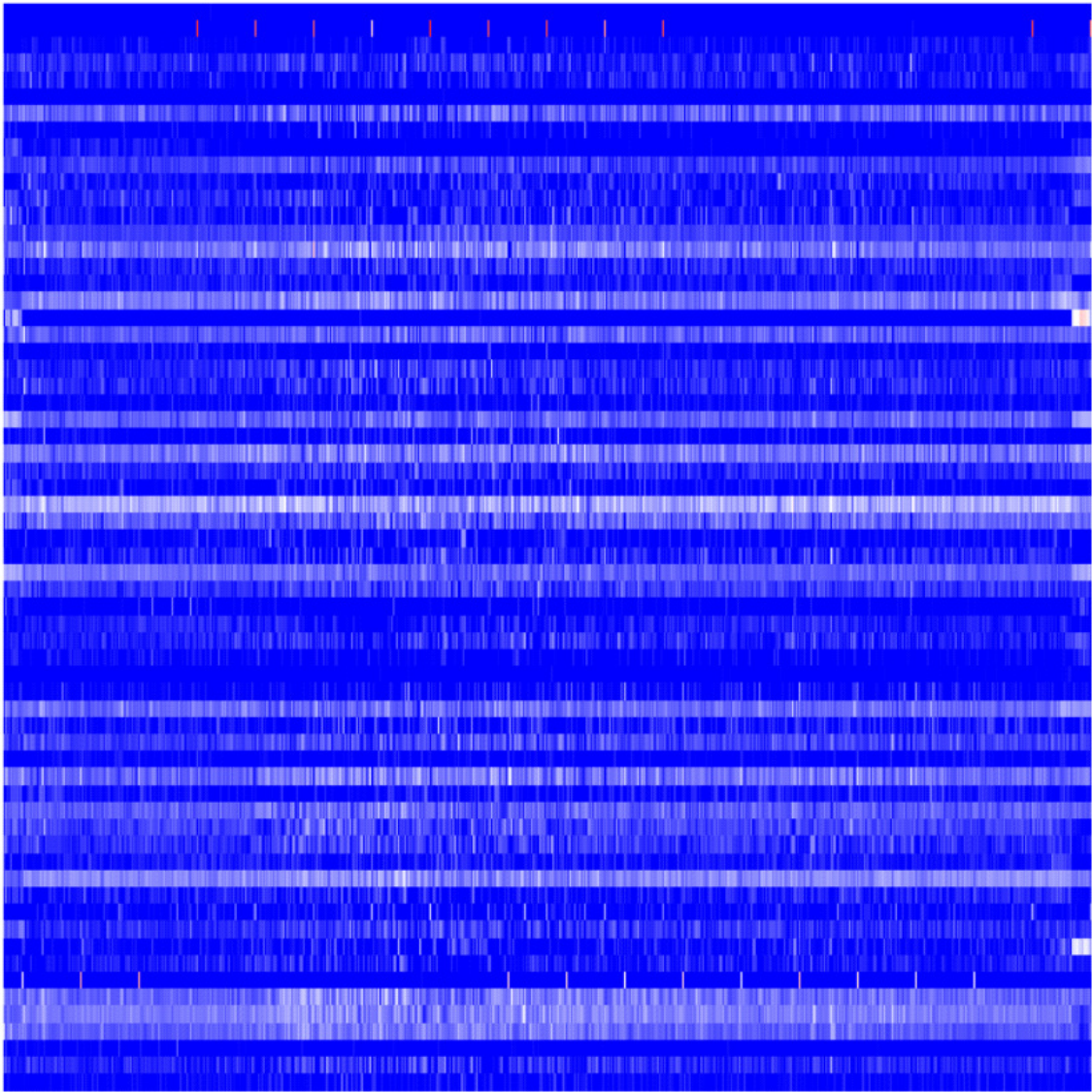}&
\includegraphics[height=18mm,clip]{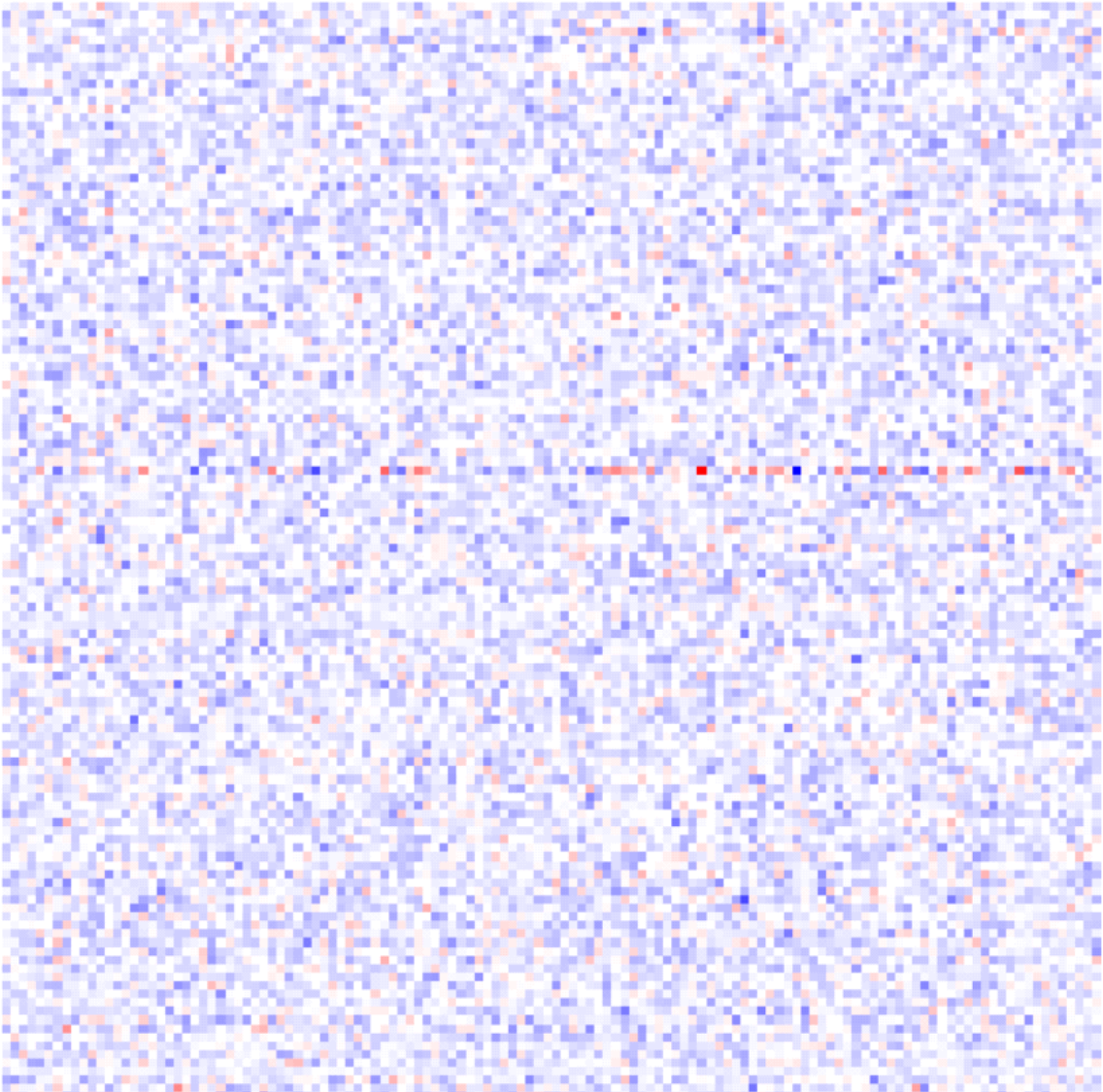}&
\hspace{-1.0em}
\includegraphics[height=18mm,clip]{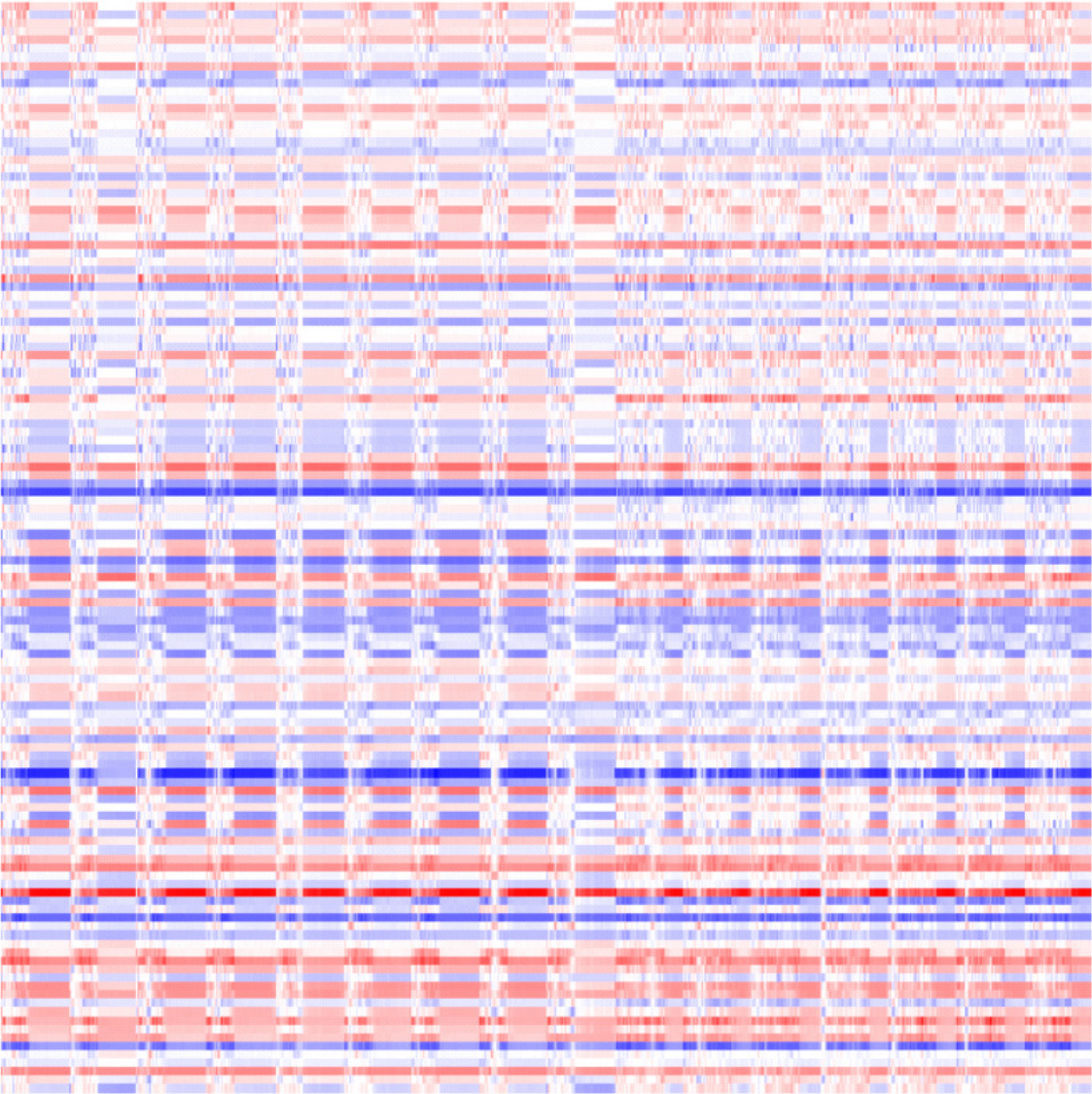}\\
Matrix $\mathbf{W}$&
\hspace{-1.0em}
Matrix $\mathbf{X}$&
Matrix $\mathbf{W}$&
\hspace{-1.0em}
Matrix $\mathbf{X}$\\
\vspace{-0.5em}
\\
\multicolumn{2}{c}{(1) ResNet50}&
\multicolumn{2}{c}{(2) BERT-Large}
\end{tabular}
\end{center}
\vspace{-0.5em}
\caption{
Heatmaps of the weight and activation matrices.
While the weight matrices exhibit irregular patterns, 
the activation matrices exhibit stripe patterns. 
After reformulating matrix multiplication as
$\mathbf{X}^{\top}\mathbf{W}^{\top}$,
this anisotropic structure is reflected in the row vectors of the
transposed activation matrix, motivating 
the proposed approach. 
}
\label{fig:mat_visualize}
\vspace{-0.0em}
\normalsize
\end{figure}

To address this problem, 
we exploit the high similarity among activation vectors
arising from \textit{anisotropy} in DNNs,
where learned embeddings are concentrated
in a small region of the representation space rather than
being uniformly distributed \cite{DBLP:conf/emnlp/Ethayarajh19,DBLP:conf/iclr/GaoHTQWL19,DBLP:conf/eacl/GodeyCS24}.
As shown in Figure~\ref{fig:mat_visualize},
activation matrices exhibit clear stripe patterns,
whereas weight matrices show no apparent patterns. 
Based on this observation, we propose \textit{SPARC
(\underline{S}imilarity-aware \underline{P}hotonic \underline{A}nalog \underline{R}esidual \underline{C}omputation)}. 
Our key idea is to reformulate matrix multiplication so that this 
structural similarity can be directly exploited,
thereby reducing the impact of quantization errors and analog photonic noise.

Instead of directly computing
$\mathbf{WX}=\mathbf{Y}$,
our approach computes the equivalent matrix multiplication
$\mathbf{X}^{\top}\mathbf{W}^{\top}=\mathbf{Y}^{\top}$,
as illustrated in Figure~\ref{fig:matrix_mult}-2.
This reformulation treats activation vectors as row vectors, allowing
each block matrix to consist of highly similar vectors due to the anisotropy.
This observation suggests that the row vectors in each block can be well approximated 
by a single vector.
To this end, we introduce a representative vector
$\mathbf r$, defined as the mean of the row vectors in each block.
Because the row vectors are highly similar,
their mean accurately represents every row vector in the block.
Let the residual vector be 
$\Delta\mathbf{a}=\mathbf{a}-\mathbf{r}$.
Then, the inner product can be decomposed as 
$\mathbf{ab} = \mathbf{rb} + \Delta\mathbf{a}\mathbf{b}$, 
where $\mathbf{rb}$ captures the dominant component of the inner product.
Since the representative vector is common within a block,
the dominant component $\mathbf r \mathbf b$
is computed only once using a high-precision digital processor,
while only the residual component $\Delta\mathbf{a}\mathbf{b}$
is obtained on the photonic accelerator.
Since the residual vectors have much smaller norms than the corresponding original vectors,
the impact of DAC/ADC quantization errors and analog photonic noise 
is substantially reduced.
Consequently, the proposed approach improves the accuracy
of photonic matrix multiplication while preserving its energy efficiency.

The conventional approach to improving the accuracy of photonic computation
is to employ higher-precision DAC/ADC \cite{DBLP:journals/jetc/DemirkiranEWEMHBRJB23}.
However, increasing the precision of DAC/ADC substantially increases
power consumption \cite{10301629,Cansu2024}.
In contrast, the proposed approach improves inference accuracy
without increasing the precision requirements of DAC/ADC,
thereby avoiding additional power consumption.
Furthermore, the proposed approach is fully compatible with existing
photonic accelerators.
As shown in Figure~\ref{fig:matrix_mult}-2, 
in our approach, the photonic device receives the residual vector
$\Delta\mathbf{a}$ instead of $\mathbf{a}$,
while the representative inner product $\mathbf{rb}$ is computed separately
on a high-precision digital processor.
Since the proposed approach preserves the original photonic
inner-product computation pipeline and requires only lightweight
preprocessing and postprocessing,
it can be integrated into existing photonic accelerators without
modifying the underlying photonic hardware.
Moreover, since the underlying photonic computation remains unchanged, 
the proposed approach is also applicable to future photonic accelerators.
Our main contributions of this paper are summarized as follows:
\begin{itemize}
\item We propose a novel approach for photonic neural
       network accelerators that exploits anisotropy in activation vectors.
       By decomposing each inner product into a representative inner product and a residual inner product,
       our approach can reduce the impact of
       DAC/ADC quantization errors and analog photonic noise.
\item We introduce a new computation paradigm that 
       preserves the existing photonic computation pipeline.
       It requires neither hardware modifications nor 
       higher-precision DAC/ADC, 
       making it readily applicable to existing and future photonic accelerators.
\item  We conduct experiments on a diverse set of DNN models and demonstrate 
       that our approach can improve inference accuracy under 
       photonic-computing constraints, 
       validating its effectiveness across different network architectures and application domains.
\end{itemize}
%
%The source code of the proposed approach will be publicly released upon acceptance,
%as a patent application for our approach is currently pending.

\section{Related Work}
\label{sec:related_work}

\paragraph{Photonic Neural Network Accelerators.}

For energy-efficient DNN inference,
photonic neural network accelerators,
which perform matrix multiplication in the optical domain,
have attracted considerable attention
\cite{DBLP:journals/access/MarinisCCA19}.
Early studies demonstrated the feasibility of optical matrix multiplication.
However, their implementations remained limited to laboratory-scale
prototypes
\cite{Shen:2017fqu,Hughes:18,10.1007/s00521-022-07243-z,Cansu2024}.
Recently, Lightmatter introduced a universal photonic accelerator
that combines photonic analog computation with adaptive quantization,
bringing photonic neural network accelerators closer to practical deployment
for large-scale DNN inference
\cite{RePEc:nat:nature:v:640:y:2025:i:8058:d:10.1038_s41586-025-08854-x}.
Despite these advances, photonic accelerators still suffer from accuracy
degradation caused by low-bit quantization errors and analog photonic noise \cite{10609496}.

\paragraph{Accuracy Enhancement.}
Existing approaches can be broadly categorized into hardware-based
and algorithm-based methods.
Hardware-based methods improve inference accuracy by employing
higher-precision DAC/ADC or hardware calibration techniques
\cite{Hailong2022,DBLP:conf/icton/DuzgolKPR23,Cansu2024}.
Although effective,
they increase hardware complexity and power consumption,
thereby diminishing the energy-efficiency benefits
of photonic computing.
Algorithm-based methods improve inference accuracy
through effective low-precision computation.
ABFP reduces quantization errors by adaptively selecting
a shared exponent for each block vector
\cite{RePEc:nat:nature:v:640:y:2025:i:8058:d:10.1038_s41586-025-08854-x,DBLP:journals/corr/abs-2205-06287}.
Calibration-based quantization determines quantization parameters
by minimizing the mean squared error
between quantized and floating-point outputs
\cite{DBLP:journals/corr/abs-2004-09602,DBLP:conf/icml/SakrDVZDK22},
but typically achieves lower inference accuracy than ABFP
\cite{DBLP:journals/corr/abs-2307-03712}.
Residue number system-based quantization exploits parallel
low-bit residue arithmetic
\cite{7738524,Cansu2024},
but requires additional residue arithmetic and conversion operations,
thereby increasing hardware complexity \cite{Cansu2024}.
Unlike existing quantization methods,
our approach improves inference accuracy
by reformulating the matrix multiplication itself,
rather than modifying the quantization procedure.
Therefore,
the proposed approach is complementary to existing quantization methods
and can be seamlessly integrated with them. 

\paragraph{Anisotropy.}
Recent studies have shown that DNN representations
are anisotropic,
where embedding vectors are concentrated in a small region
of the representation space
rather than being uniformly distributed
\cite{DBLP:conf/emnlp/Ethayarajh19,DBLP:conf/iclr/GaoHTQWL19,DBLP:conf/eacl/GodeyCS24}.
These studies primarily exploit anisotropy
for representation analysis and embedding enhancement,
whereas we leverage it to improve
inference accuracy of photonic neural network accelerators.
Specifically, we construct a representative vector for each block
of similar activation vectors,
thereby reducing the impact of quantization errors and analog photonic noise
without modifying the underlying hardware.

\section{Preliminaries}
\label{sec:preliminary}

Given a weight matrix $\mathbf W$ and an activation matrix $\mathbf X$,
DNN inference computes the matrix product
$\mathbf{WX}=\mathbf{Y}$.
To improve computational efficiency,
the universal photonic accelerator maps $\mathbf{W}$, $\mathbf{X}$, and $\mathbf{Y}$
to matrices $\mathbf{A}$, $\mathbf{B}$, and $\mathbf{C}$,
respectively, 
and decomposes the matrix multiplication 
into vector inner products \cite{RePEc:nat:nature:v:640:y:2025:i:8058:d:10.1038_s41586-025-08854-x}.
As illustrated in Figure~\ref{fig:matrix_mult}-1,
matrix $\mathbf{A}$ is partitioned into $n\times n$ block matrices,
while matrix $\mathbf{B}$ is partitioned into
corresponding block column vectors of length $n$.
Let $\mathbf{a}$ be a row vector from a block of $\mathbf{A}$,
and let $\mathbf{b}$ be the corresponding block column vector.
The photonic device computes their inner product,
$\mathbf{ab}=c$.

Before DAC conversion,
the vectors are normalized and quantized using ABFP
\cite{DBLP:journals/corr/abs-2205-06287}.
The normalized vector $\hat{\mathbf{a}}$ corresponding to
$\mathbf{a}$ is given by
\begin{equation}
\label{eq:normalize_vec_a}
\hat{\mathbf{a}} = \mathbf{a}/s^{\mathbf{a}},
\end{equation}
where $s^{\mathbf{a}} = \max_{1 \le i \le n}( | a[i] | )$ is 
the scaling factor stored in bfloat16 format \cite{DBLP:journals/corr/AbadiABBCCCDDDG16}.
Each element of $\hat{\mathbf{a}}$ lies in $[-1,1]$.
Similarly, vector $\mathbf{b}$ is normalized as $\hat{\mathbf{b}} = \mathbf{b}/s^{\mathbf{b}}$
where $s^{\mathbf{b}}$ is the scaling factor for $\mathbf{b}$. 

The normalized vectors are then uniformly quantized.
The quantized vector of $\hat{\mathbf{a}}$,
denoted by $\mathbf{a}^{q}$,
is computed using the quantization function $Q(\cdot)$:
\begin{equation}
\label{eq:quantize_vec_a}
\textstyle \mathbf{a}^q = Q(\hat{\mathbf{a}}, \delta_{\mathbf{A}}, \tau_{\mathbf{A}}) = \mathrm{clamp}  (  \lfloor \frac{\hat{\mathbf{a}}}{\delta_{\mathbf{A}}} \rceil \delta_{\mathbf{A}}, \tau_{\mathbf{A}}),
\end{equation}
where $\mathrm{clamp}(\cdot)$ constrains each element to the range
$[-\tau_{\mathbf{A}}, \tau_{\mathbf{A}}]$ and is defined as
\begin{equation}
\label{eq:clamp}
\mathrm{clamp}(\hat{\mathbf{a}}, \tau_{\mathbf{A}}) = \max(\min(\hat{\mathbf{a}}, \tau_{\mathbf{A}}), -\tau_{\mathbf{A}}). 
\end{equation}
Since $\hat{\mathbf a}$ lies in $[-1,1]$,
$\tau_{\mathbf{A}}$ is set to $1$.
In Equation~(\ref{eq:quantize_vec_a}), 
$\delta_{\mathbf{A}}$ denotes the quantization step size,
defined as
$\delta_{\mathbf{A}} \! = \! 1/(2^{b_{\mathbf{A}} \! - \! 1} \! - \! 1)$ 
where $b_{\mathbf{A}}$ denotes the quantization bit width.
The operator $\lfloor\cdot\rceil$ denotes round-half-to-even rounding. 
Similarly, vector $\mathbf{b}$ is quantized as
$\mathbf{b}^q =Q(\hat{\mathbf{b}}, \delta_{\mathbf{B}}, \tau_{\mathbf{B}})$
by setting $\tau_{\mathbf{B}}=1$ and 
$\delta_{\mathbf{B}} \! = \! 1/(2^{b_{\mathbf{B}} \! - \! 1} \! - \! 1)$
where $b_{\mathbf{B}}$ is the quantization bit width. 

The quantized vectors are converted into analog signals by the DAC,
after which
the photonic device computes
\begin{equation}
\label{eq:c_q}
c = s^{\mathbf{a}} s^{\mathbf{b}} Q(\mathbf{a}^q \mathbf{b}^q, n \delta_{\mathbf{C}}, \tau_{\mathbf{C}}).
\end{equation}
In Equation~(\ref{eq:c_q}),
$\tau_{\mathbf{C}} \! = \! n$ and $\delta_{\mathbf{C}} \! = \! 1/(2^{b_{\mathbf{C}} - 1} \! - \! 1)$ 
where $b_{\mathbf{C}}$ denotes the ADC quantization bit width. 
The inner product $\mathbf{a}^q \mathbf{b}^q$ is computed in the analog domain,
whereas the scaling operation $s^{\mathbf{a}}s^{\mathbf{b}}$ 
is performed digitally after ADC conversion.
Since $\mathbf{a}^{q}$ and $\mathbf{b}^{q}$ are normalized,
their inner product lies in $[-n,n]$. 
Therefore, $\delta_{\mathbf{C}}$ and $\tau_{\mathbf{C}}$ are scaled by $n$. 
The output is stored in bfloat16 format after ADC quantization.

As the block length $n$ increases,
the dynamic range of the accumulated inner product also increases.
With a fixed ADC precision,
this wider dynamic range reduces the effective ADC resolution
for small-magnitude values.
To improve the effective ADC resolution without increasing the ADC precision,
the photonic analog output is amplified by a gain factor $G \ge 1$
before analog-to-digital conversion:
\begin{equation}
\label{eq:c_q_gain}
c = s^{\mathbf{a}} s^{\mathbf{b}} Q(G \mathbf{a}^q \mathbf{b}^q, n \delta_{\mathbf{C}}, \tau_{\mathbf{C}})/G.
\end{equation}
Although amplification reduces the effective ADC resolution
for large values and introduces extra power consumption \cite{8110629},
it improves the resolution for small values.

However, photonic matrix multiplication suffers from
quantization errors and analog photonic noise.
The input vectors are quantized before DAC conversion,
whereas the analog outputs are quantized by the ADC,
introducing numerical errors at both stages.
More importantly, the analog photonic computation itself is affected by
analog photonic noise caused by optical signal distortions
and hardware imperfections
\cite{RePEc:nat:nature:v:640:y:2025:i:8058:d:10.1038_s41586-025-08854-x}.
Such noise cannot be completely eliminated because
it inevitably arises within the analog computation itself. 
Let $\varepsilon$ denote additive analog photonic noise.
The resulting inner product is computed as
\begin{equation}
\label{eq:c_q_noise}
c = s^{\mathbf{a}} s^{\mathbf{b}} Q(G \mathbf{a}^q \mathbf{b}^q + \varepsilon, n \delta_{\mathbf{C}}, \tau_{\mathbf{C}})/G,
\end{equation}
where $\varepsilon$ is modeled as additive zero-mean Gaussian noise with variance $\sigma^2$
\cite{DBLP:conf/dac/RekhiZNLVWKDG19,DBLP:journals/corr/abs-2205-06287}.

\begin{algorithm}[t]
%\scriptsize
%\footnotesize
\small
\caption{Original Photonic Matrix Multiplication}
\label{alg:original}
%\algsetup{linenosize=\scriptsize}
\begin{algorithmic}[1]
	\Input  weight matrix $\mathbf{W}$, activation matrix $\mathbf{X}$, block length $n$, 
                bit width $b_{\mathbf{A}}$, bit width $b_{\mathbf{B}}$, bit width $b_{\mathbf{C}}$, and gain factor $G$
        \Output matrix $\mathbf{Y}$
        
        \State $\mathbf{A} = \mathbf{W}$, $\mathbf{B} = \mathbf{X}$, $\mathbf{C} = \mathbf{0}$; 
        \State $\delta_{\mathbf{A}} \! = \! 1/(2^{b_{\mathbf{A}} - 1} \! - \! 1)$, $\delta_{\mathbf{B}} \! = \! 1/(2^{b_{\mathbf{B}} - 1} \! - \! 1)$, $\delta_{\mathbf{C}} \! = \! 1/(2^{b_{\mathbf{C}} - 1} \! - \! 1)$;
        \State partition $\mathbf{A}$ into $n \times n$ blocks;
        \State partition $\mathbf{B}$ into column blocks of length $n$;
        
        \For{each block}
        	\For{each row vector $\mathbf{a}$ in the block matrix}
        		\State compute $\mathbf{a}^q = \mathrm{ABFP} (\mathbf{a}, n, \delta_{\mathbf{A}})$ using Algorithm~\ref{alg:abfp};
        	\EndFor
        	\For{each corresponding column vector $\mathbf{b}$}
                	\State compute $\mathbf{b}^q = \mathrm{ABFP} (\mathbf{b}, n, \delta_{\mathbf{B}})$ using Algorithm~\ref{alg:abfp};
                	\For{each quantized vector $\mathbf{a}^q$ in the block matrix}
        			\State compute $c = s^{\mathbf{a}} s^{\mathbf{b}} Q(G \mathbf{a}^q \mathbf{b}^q, n \delta_{\mathbf{C}}, n)/G$;
                                \State accumulate $c$ into the corresponding element of $\mathbf{C}$;
                        \EndFor
                \EndFor
        \EndFor
        
        \State $\mathbf{Y} = \mathbf{C}$;        
\end{algorithmic}
\normalfont
\end{algorithm}

\begin{algorithm}[t]
%\scriptsize
%\footnotesize
\small
\caption{ABFP Quantization}
\label{alg:abfp}
%\algsetup{linenosize=\scriptsize}
\begin{algorithmic}[1]
	\Input  vector $\mathbf{a}$, block length $n$, and quantization step size $\delta_{\mathbf{A}}$
        \Output quantized vector $\mathbf{a}^q$
        
        \State $s^{\mathbf{a}} = \max_{1 \le i \le n}( | a[i] | )$;
        \State $\hat{\mathbf{a}} = \mathbf{a}/s^{\mathbf{a}}$;
        \State $\mathbf{a}^q = Q(\hat{\mathbf{a}}, \delta_{\mathbf{A}}, 1)$;        
\end{algorithmic}
\normalfont
\end{algorithm}

Algorithm~\ref{alg:original} outlines the original photonic matrix-multiplication procedure.
To compute the matrix product $\mathbf{WX}=\mathbf{Y}$,
it initializes $\mathbf{A}$ and $\mathbf{B}$ from
$\mathbf{W}$ and $\mathbf{X}$, respectively,
and initializes $\mathbf{C}$ to the zero matrix (line~1).
It then partitions $\mathbf A$ into $n\times n$ blocks 
and $\mathbf B$ into column blocks of length $n$ (lines~3--4).
For each block, the row and column vectors are quantized 
using Algorithm~\ref{alg:abfp} (lines~6--9).
The photonic device computes the inner products of the quantized vectors
and accumulates the resulting values into $\mathbf{C}$
to produce the output matrix (lines~10--12).
The computational and memory costs of  original photonic matrix-multiplication approach are summarized below.
\begin{lemma}
\label{lem:ori_cost}
Let $\mathbf{A}\in\mathbb{R}^{N\times D}$ and $\mathbf{B}\in\mathbb{R}^{D\times M}$.
The original photonic matrix-multiplication approach requires $O(ND+MD+NMD/n)$
time and $O(ND+MD+NM)$ space.
\end{lemma}
\begin{proof}
Quantizing matrices $\mathbf A$ and $\mathbf B$ requires
$O(ND)$ and $O(MD)$ time, respectively.
The photonic device computes all inner products, 
while the digital processor performs one scaling operation per inner product.
Since the number of inner products is $NMD/n$, the total computational cost is
$O(ND+MD+NMD/n)$.
Memory consumption is dominated by storing $\mathbf{A}$, $\mathbf{B}$, and
$\mathbf{C}$ with sizes $ND$, $MD$, and $NM$, respectively.
The storage required for the scaling factors is negligible
compared with that required for the matrices.
Therefore, the memory complexity is $O(ND+MD+NM)$.
\end{proof}
%The proof of this lemma is shown in the appendix.
Compared with conventional digital matrix multiplication,
which requires $O(NMD)$ time and $O(ND+MD+NM)$ space,
the universal photonic accelerator reduces the computational complexity to
$O(ND+MD+NMD/n)$ without increasing the memory complexity.
This improvement, however, comes at the expense of inference accuracy due to
DAC/ADC quantization errors and analog photonic noise.
The next section presents our approach
to mitigating these degradations. 
%without modifying the underlying photonic hardware.

\section{Proposed Approach}
\label{sec:proposed_method}

\begin{figure}[t]
\begin{center}
\small
%\footnotesize
\begin{tabular}{cc}
\begin{minipage}{0.6\hsize}
\begin{tabular}{cc}
\includegraphics[height=24mm,clip]{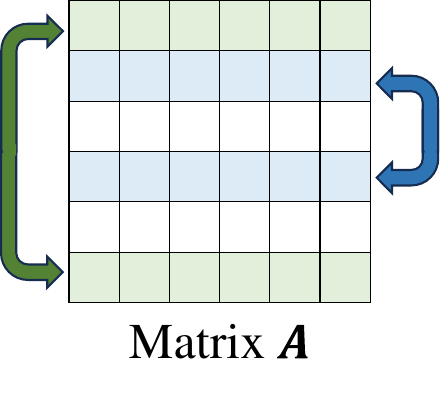}&
\hspace{-1.0em}
\includegraphics[height=24mm,clip]{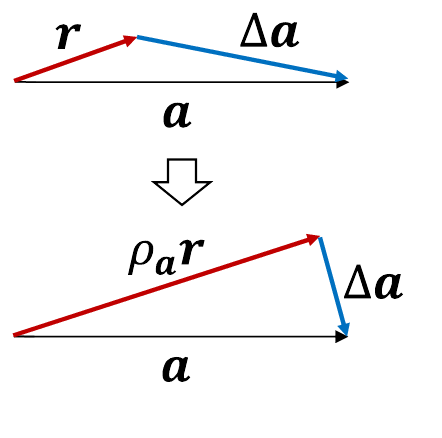}\\
(1) Reordering \vspace{-0.8em}&
\hspace{-2.0em}
(2) Scaling \vspace{-0.8em}
\end{tabular}
\caption{Proposed approach.}
\label{fig:pro_outline}
\end{minipage}
\begin{minipage}{0.39\hsize}
\begin{center}
\includegraphics[height=24mm,clip]{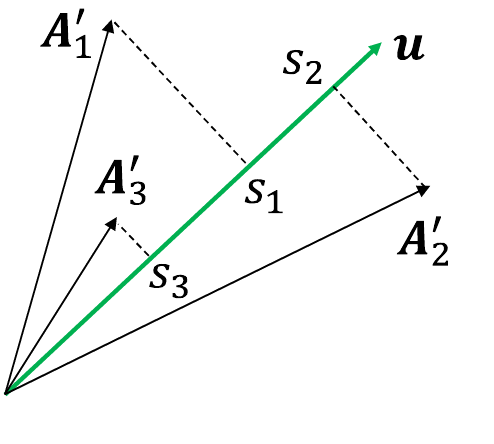}
\end{center}
\vspace{-1.9em}
\caption{Vector projection.}
\label{fig:vec_projection}
\end{minipage}
\end{tabular}
\normalsize
\end{center}
\vspace{-0.0em}
\end{figure}

The proposed approach improves the inference accuracy of photonic matrix multiplication 
by exploiting the anisotropic structure of activation vectors in DNNs.
After reformulating the matrix multiplication as $\mathbf{X}^{\top}\mathbf{W}^{\top}$, 
the activation vectors correspond to the row vectors of matrix $\mathbf{A}$,
allowing each block to contain similar activation vectors.
For each block, we compute a representative vector
$\mathbf{r}=\frac{1}{n}\sum_{i=1}^{n}\mathbf{a}_i$,
where $\mathbf{a}_i$ denotes the $i$-th row vector.
Given a row vector $\mathbf{a}$ in the block,
its residual vector is defined as $\Delta\mathbf{a}=\mathbf{a}-\mathbf{r}$,
and thus the original inner product is rewritten as 
$\mathbf{ab} = \mathbf{rb} + \Delta\mathbf{a}\mathbf{b}$. 
The dominant component $\mathbf{rb}$ is computed on a high-precision digital processor, 
whereas the residual component $\Delta\mathbf{a}\mathbf{b}$ is computed on the photonic hardware.
As shown in Figure~\ref{fig:pro_outline}, 
we further improve the inference accuracy using the following two methods:
\begin{itemize}
\item We reorder the row vectors of matrix $\mathbf{A}$ so that
adjacent vectors become more similar, enabling each representative
vector to effectively approximate its block.
\item We scale the representative vector before computing the residual
vector, thereby reducing the residual norm 
processed on the photonic device.
\end{itemize}
The following subsections describe these two methods.

\subsection{Row Vector Reordering}

Before reordering, we center the columns of matrix $\mathbf{A}$.
This step removes the common offset shared across row vectors,
improving the effectiveness of the subsequent reordering.
Let the column-wise mean vector be
\begin{equation}
\textstyle \mathbf{m} = \frac{1}{N} \sum_{i=1}^{N} \mathbf{A}_i,
\end{equation}
where $\mathbf{A}_i$ denotes the $i$-th row vector in matrix $\mathbf{A}$.
If $\mathbf{A}_{i}^\prime$ is the mean-centered row vector of $\mathbf{A}_i$, 
it is given as $\mathbf{A}_{i}^\prime=\mathbf{A}_{i} - \mathbf{m}$. 
Since matrix $\mathbf{A}$ has size $N\times D$,
mean-centering all row vectors requires $O(ND)$ time.
Let $\mathbf{B}_j$ denote the $j$-th column vector of matrix
$\mathbf{B}$.
If $\mu_j = \mathbf{mB}_j$, 
each inner product in $\mathbf{AB}=\mathbf{C}$
is rewritten as
\begin{equation}
\label{eq:nomal_ab_inner}
\textstyle \mathbf{A}_{i} \mathbf{B}_{j} = 
\mathbf{A}_{i}^\prime \mathbf{B}_{j} + \mathbf{m} \mathbf{B}_{j} = \mathbf{A}_{i}^\prime \mathbf{B}_{j} + \mu_j. 
\end{equation}
Equation~(\ref{eq:nomal_ab_inner}) decomposes each inner product into
a mean-centered component and a shared offset.
Since $\mu_j$ is shared by all row vectors,
it is computed once for each column on the digital processor
in $O(MD)$ time for the matrix multiplication,
whereas the computationally intensive term $\mathbf{A}_i^\prime \mathbf{B}_j$
is obtained on the photonic device. 
We store $\mu_j$ in bfloat16 format.

After mean-centering, 
we reorder the row vectors so that 
adjacent row vectors become more similar.
In the reordering, 
we project each mean-centered row vector onto a unit vector $\mathbf{u}$ satisfying
$\|\mathbf{u}\|^2=1$ and 
sort each vector according to its projection score $s_i$,
which is given as follows: 
\begin{equation}
s_i=\mathbf{A}_{i}^\prime \mathbf{u}^\top.
\end{equation}
As illustrated in Figure~\ref{fig:vec_projection}, 
the vectors are projected onto a single unit vector. 
If several row vectors have similar projection scores,
the projection fails to preserve their relative differences, 
resulting in an ineffective ordering.

We therefore seek the unit vector that maximizes the variance of 
the projection scores.
Let $V(\mathbf{u})$ denote the variance of the projection scores produced 
by unit vector $\mathbf{u}$.
Since each row vector is mean-centered as $\mathbf{A}_{i}^\prime=\mathbf{A}_{i} - \mathbf{m}$, 
we have the following equation: 
\begin{equation}
\textstyle \sum_{i=1}^N A_{i}^\prime[j] = \sum_{i=1}^N (A_{i}[j] - m[i]) = 0.
\end{equation}
Using this property, the variance of the projection scores becomes
\begin{equation}
\label{eq:Vari_u}
\begin{split}
\!\! V(\mathbf{u}) & = \textstyle \frac{1}{N} \sum_{i=1}^N (\mathbf{A}_{i}^\prime \mathbf{u}^\top)^2 -
                           (\frac{1}{N} \sum_{i=1}^N \mathbf{A}_{i}^\prime \mathbf{u}^\top)^2 \\
&= \textstyle \frac{1}{N} \mathbf{u} \mathbf{A}^\top \mathbf{A} \mathbf{u}^\top - (\frac{1}{N} \sum_{i=1}^N \sum_{j=1}^D A_{i}^\prime[j] {u}[j])^2 \!\! \\
&= \textstyle \frac{1}{N} \mathbf{u} \mathbf{A}^\top \mathbf{A} \mathbf{u}^\top.
\end{split}
\end{equation}
To maximize the variance under the constraint $\|\mathbf{u}\|^2=1$,
we consider the following Lagrangian:
\begin{equation}
\textstyle L(\mathbf{u},\lambda) = \frac{1}{N} \mathbf{u} \mathbf{A}^\top \mathbf{A} \mathbf{u}^\top - \lambda(\mathbf{u} \mathbf{u}^\top -1),
\end{equation}
where $\lambda$ denotes Lagrange multiplier.
Differentiating $L(\mathbf{u},\lambda)$ with respect to $\mathbf{u}^{\top}$
and setting the derivative to zero yields 
\begin{equation}
\label{eq:div_Vari_u}
\textstyle \frac{1}{N} \mathbf{A}^\top \mathbf{A} \mathbf{u}^\top = \lambda \mathbf{u}^\top.
\end{equation}
Equation~(\ref{eq:div_Vari_u}) indicates that $\mathbf{u}^{\top}$
is an eigenvector of matrix $\frac{1}{N}\mathbf{A}^{\top}\mathbf{A}$,
and that $\lambda$ is the corresponding eigenvalue.
Substituting Equation~(\ref{eq:div_Vari_u}) into Equation~(\ref{eq:Vari_u}) gives
\begin{equation}
\label{eq:eigen_u}
\textstyle V(\mathbf{u}) =  \mathbf{u} \lambda \mathbf{u}^\top = \lambda \| \mathbf{u} \|^2 = \lambda.
\end{equation}
Therefore, the variance of the projection scores is equal to
the corresponding eigenvalue of
$\frac{1}{N}\mathbf{A}^{\top}\mathbf{A}$.
Since the row vectors of matrix $\mathbf{A}$ are mean-centered,
$\frac{1}{N}\mathbf{A}^{\top}\mathbf{A}$
is the covariance matrix of $\mathbf{A}$. 
Hence, the optimal projection direction 
is given by the first principal component
of matrix $\mathbf A$.
Our method adopts the CC-IPCA algorithm \cite{DBLP:journals/pami/WengZH03} 
to efficiently compute the first principal component as follows:
\begin{equation}
\label{eq:CC-IPCA}
\begin{split}
&\!\!\!\! \mathbf{u} =  \textstyle \, \mathbf{u}_N, \; \text{where} \; \mathbf{u}_i = \mathbf{u}^\prime_i / \| \mathbf{u}^\prime_i \|, \\
&\!\!\!\! \textstyle \mathbf{u}^\prime_i = 
\begin{cases}
\textstyle \mathbf{A}_{i}^\prime \quad \text{if } i=1 \\ 
\textstyle \frac{i-1}{i} \mathbf{u}_{i-1} + \frac{1}{i} \mathbf{u}_{i-1} (\mathbf{A}_{i}^\prime)^\top \mathbf{A}_{i}^\prime $\;\;$ \text{if } i = 2, \ldots, N.
\end{cases}
\end{split}
\end{equation}
Since matrix $\mathbf{A}$ is $N\times D$,
Equation~(\ref{eq:CC-IPCA}) computes unit vector $\mathbf{u}$ in $O(ND)$ time.
Since $s_i=\mathbf{A}_{i}^{\prime}\mathbf{u}^{\top}$,
the projection scores can also be computed in $O(ND)$ time.

We perform bucket sort to efficiently reorder the row vectors \cite{Cormen2022}. 
As shown in Figure~\ref{fig:matrix_mult}, each block contains $n$ row vectors, 
resulting in $N/n$ blocks in each column in matrix $\mathbf{A}$. 
Therefore, if $m=N/n$ is the number of buckets, 
we quantize each projection score into a bucket index as
\begin{equation}
\label{eq:bucket}
\textstyle s_i^\prime = \min(\lfloor m\frac{s_i-s_{\min}}{s_{\max}-s_{\min}}\rfloor,m-1),
\end{equation}
where $s_{\min} \! = \! \min_{1 \le i \le N} (s_i)$ and $s_{\max} \! = \! \max_{1 \le i \le N} (s_i)$. 
The operator $\lfloor \cdot \rfloor$ denotes floor rounding.
Equation~(\ref{eq:bucket}) quantizes each projection score to 
an integer in the range $[0,m-1]$.
The reordered matrix is then obtained by concatenating the row vectors stored 
in the buckets in ascending bucket order.
Both score quantization and bucket assignment require $O(N)$ time, 
yielding an overall sorting cost of $O(N)$.

Let $\mathbf{A}^{*}$ denote the reordered matrix, and let $\mathbf{R}$ 
be the $N\times N$ reordering matrix representing the row reordering of $\mathbf{A}$.
Each row and each column of $\mathbf{R}$ contains exactly one element equal to
$1$, while all remaining elements are zero.
Since $\mathbf{A}^* = \mathbf{R} \mathbf{A}$ and $\mathbf{R}^{-1}=\mathbf{R}^\top$, 
matrix multiplication $\mathbf{A}\mathbf{B}$ can be rewritten as
\begin{equation}
\label{eq:pemutate_ABC}
\textstyle \mathbf{A} \mathbf{B} = \mathbf{R}^{\top} \mathbf{A}^* \mathbf{B}.
\end{equation}
Thus, photonic matrix multiplication is performed on the reordered matrix $\mathbf{A}^{*}$, 
after which the original row order is restored by applying the inverse reordering.

\subsection{Scaling Representative Vector}

The representative-vector decomposition
$\mathbf{ab}=\mathbf{rb}+\Delta\mathbf{a}\mathbf{b}$
reduces photonic computation errors, since
the dominant component $\mathbf{rb}$ is computed on a high-precision
digital processor.
We scale the representative vector to further reduce the residual norm.

Let $\rho_{\mathbf{a}}$ denote the scaling parameter associated with
vector $\mathbf{a}$. 
After scaling the representative vector, the residual vector is given as
$\Delta \mathbf{a} = \mathbf{a} - \rho_{\mathbf{a}} \mathbf{r}$. 
For simplicity, we assume that the representative vector is normalized,
i.e., $\| \mathbf{r} \|^2 =1$. 
Then, the squared norm of the residual vector is given as follows:
\begin{equation}
\textstyle \| \Delta \mathbf{a} \|^2 \! = \! 
(\mathbf{a} \! - \! \rho_{\mathbf{a}} \mathbf{r}) (\mathbf{a} \! - \! \rho_{\mathbf{a}} \mathbf{r})^\top
\! = \! \| \mathbf{a} \|^2 \! - \! 2 \rho_{\mathbf{a}} \mathbf{a} \mathbf{r}^\top \! + \! \rho_{\mathbf{a}}^2. 
\end{equation}
The scaling parameter is obtained by minimizing
the norm of $\Delta \mathbf{a}$.
Differentiating $\| \Delta \mathbf{a} \|^2$ with respect to $\rho_{\mathbf{a}}$ yields
\begin{equation}
\textstyle \frac{\partial \| \Delta \mathbf{a} \|^2}{\partial \rho_{\mathbf{a}}} 
= -2 \mathbf{a} \mathbf{r}^\top + 2 \rho_{\mathbf{a}}. 
\end{equation}
Since the norm of $\Delta \mathbf{a}$ is minimized when this derivative becomes zero, 
the scaling parameter is given as follows: 
\begin{equation}
\label{eq:scale_parameter}
\textstyle \rho_{\mathbf{a}} = \mathbf{a} \mathbf{r}^\top.
\end{equation}
Since the block length is $n$, computing $\rho_{\mathbf{a}}$
requires $O(n)$ time.
The scaling parameters are stored in bfloat16 format.
Using this scaling parameter,
vector $\mathbf{a}$ can be decomposed as 
$\mathbf{a} = \rho_{\mathbf{a}} \mathbf{r} + \Delta \mathbf{a}$, 
and the inner product $\mathbf{ab}$ becomes as follows:
\begin{equation}
\label{eq:scale_rep_ab}
\textstyle \mathbf{ab} = \rho_{\mathbf{a}} \mathbf{r} \mathbf{b} + \Delta \mathbf{a} \mathbf{b}.
\end{equation}

To further improve the computational accuracy of Equation~(\ref{eq:scale_rep_ab}), 
we introduce a scaling parameter for vector $\mathbf{b}$.
Let $\rho_{\mathbf{b}}$ denote the scaling parameter associated with vector $\mathbf{b}$,
and let $\Delta \mathbf{b}$ denote the corresponding residual vector.
Then, the residual vector of $\mathbf{b}$ is given as 
$\Delta \mathbf{b} = \mathbf{b} - \rho_{\mathbf{b}} \mathbf{r}^\top$. 
Therefore, the squared norm of $\Delta\mathbf{b}$ is
\begin{equation}
\textstyle \!\!\! \| \Delta \mathbf{b} \|^2 \!_{\!} = \! 
(\mathbf{b} \! - \! \rho_{\mathbf{b}} \mathbf{r}^{\!\top})^{\!\top} \! (\mathbf{b} \! - \! \rho_{\mathbf{b}} \mathbf{r}^{\!\top})
\! = \! \| \mathbf{b} \|^2 \!_{\!} - \! 2 \rho_{\mathbf{b}} \mathbf{r} \mathbf{b} \! + \! \rho_{\mathbf{b}}^2. \!\!\! 
\end{equation}
Similarly, by differentiating $\| \Delta \mathbf{b} \|^2$
with respect to $\rho_{\mathbf{b}}$
and setting the derivative to zero, 
we can compute the scaling parameter in $O(n)$ time, which is given as follows:
\begin{equation}
\label{eq:scale_parameter_b}
\textstyle \rho_{\mathbf{b}} = \mathbf{r}\mathbf{b}. 
\end{equation}
Since $\mathbf{b} \! = \! \rho_{\mathbf{b}} \mathbf{r}^\top \! + \! \Delta \mathbf{b}$, 
inner product $\mathbf{ab}$ becomes 
\begin{equation}
\label{eq:scale_rep_ab_b}
\textstyle \mathbf{ab} = 
\rho_{\mathbf{a}} \rho_{\mathbf{b}} + 
\rho_{\mathbf{a}} \mathbf{r} \Delta \mathbf{b}+
\rho_{\mathbf{b}} \Delta \mathbf{a} \mathbf{r}^\top+
\Delta \mathbf{a} \Delta \mathbf{b}. 
\end{equation}
From the definitions of $\rho_{\mathbf{a}}$ and $\rho_{\mathbf{b}}$,
we have 
\begin{align}
&\textstyle \mathbf{r} \Delta \mathbf{b} \! = \! \mathbf{r}(\mathbf{b} \! - \! \rho_{\mathbf{b}} \mathbf{r}^\top) \! = \! \mathbf{r}\mathbf{b} \! - \! \| \mathbf{r} \|^2 \mathbf{r}\mathbf{b} \! = \! 0,\\
&\textstyle \Delta \mathbf{a} \mathbf{r}^\top \! = \! (\mathbf{a} \! - \! \rho_{\mathbf{a}} \mathbf{r}) \mathbf{r}^\top \! = \! \mathbf{a}\mathbf{r}^\top \! - \! \| \mathbf{r} \|^2 \mathbf{a}\mathbf{r}^\top \! = \! 0.
\end{align}
Consequently, Equation~(\ref{eq:scale_rep_ab_b}) can be simplified as
\begin{equation}
\label{eq:scale_rep_ab_b_re}
\textstyle \mathbf{ab} = 
\rho_{\mathbf{a}} \rho_{\mathbf{b}} + \Delta \mathbf{a} \Delta \mathbf{b}. 
\end{equation}
Without representative-vector scaling, 
since $\mathbf{ab}=\mathbf{rb}+\Delta\mathbf{a}\mathbf{b}$, 
the digital processor computes $\mathbf{rb}$, 
whereas the photonic device computes $\Delta\mathbf{a}\mathbf{b}$, 
as shown in Figure~\ref{fig:matrix_mult}-2. 
With representative-vector scaling, from Equation~(\ref{eq:scale_rep_ab_b_re}),
the digital processor computes the scalar product $\rho_{\mathbf{a}}\rho_{\mathbf{b}}$ in $O(1)$ time, 
while the photonic device computes $\Delta\mathbf{a}\Delta\mathbf{b}$.
Since the scaling parameters minimize the norms of the residual vectors, 
the photonic computation operates on substantially smaller vectors, 
thereby reducing the impact of DAC/ADC quantization errors and analog photonic noise.

\begin{algorithm}[t]
%\scriptsize
%\footnotesize
\small
\caption{Proposed Photonic Matrix Multiplication}
\label{alg:proposed}
%\algsetup{linenosize=\scriptsize}
\begin{algorithmic}[1]
	\Input  weight matrix $\mathbf{W}$, activation matrix $\mathbf{X}$, block length $n$, 
                bit width $b_{\mathbf{A}}$, bit width $b_{\mathbf{B}}$, bit width $b_{\mathbf{C}}$, and gain factor $G$
        \Output matrix $\mathbf{Y}$
        
        \State $\mathbf{A} = \mathbf{X}^\top$, $\mathbf{B} = \mathbf{W}^\top$, $\mathbf{C} = \mathbf{0}$; 
        \State $\delta_{\mathbf{A}} \! = \! 1/(2^{b_{\mathbf{A}} - 1} \! - \! 1)$, $\delta_{\mathbf{B}} \! = \! 1/(2^{b_{\mathbf{B}} - 1} \! - \! 1)$, $\delta_{\mathbf{C}} \! = \! 1/(2^{b_{\mathbf{C}} - 1} \! - \! 1)$;
        
        \State construct the reordered matrix $\mathbf{A}^*$ using Algorithm~\ref{alg:reordering};
        
        \State partition $\mathbf{A}^*$ into $n \times n$ blocks;
        \State partition $\mathbf{B}$ into column blocks of length $n$;
        
        \For{each block}
        	\State compute the representative vector $\mathbf{r}=\frac{1}{n}\sum_{i=1}^{n}\mathbf{a}_i$;
                \State normalize the representative vector $\mathbf{r}=\mathbf{r}/\| \mathbf{r} \|$;
        	\For{each row vector $\mathbf{a}$ in the block matrix}
                	\State compute $\rho_{\mathbf{a}} = \mathbf{a} \mathbf{r}^\top$;
                        \State compute $\Delta \mathbf{a} = \mathbf{a} - \rho_{\mathbf{a}} \mathbf{r}$;
        		\State compute $\mathbf{a}^q = \mathrm{ABFP} (\Delta \mathbf{a}, n, \delta_{\mathbf{A}})$ using Algorithm~\ref{alg:abfp};
        	\EndFor
        	\For{each corresponding column vector $\mathbf{b}$}
                	\State compute $\rho_{\mathbf{b}} = \mathbf{r}\mathbf{b}$;
                        \State compute $\Delta \mathbf{b} = \mathbf{b} - \rho_{\mathbf{b}} \mathbf{r}^\top$;
                	\State compute $\mathbf{b}^q = \mathrm{ABFP} (\Delta \mathbf{b}, n, \delta_{\mathbf{B}})$ using Algorithm~\ref{alg:abfp};
                	\For{each row vector $\mathbf{a}^q$ in the block matrix}
        			\State compute $c = s^{\mathbf{a}} s^{\mathbf{b}} Q(G \mathbf{a}^q \mathbf{b}^q, n \delta_{\mathbf{C}}, n)/G$;
                                \State update $c$ as $c = \rho_{\mathbf{a}}\rho_{\mathbf{b}} +c$;
                                \State accumulate $c$ into the corresponding element of $\mathbf C$;
                        \EndFor
                \EndFor
        \EndFor
        
        \For{$j=1$ to $M$}
        	\State $\mu_j = \mathbf{mB}_j$;
        \EndFor
        
        \For{$i=1$ to $N$}
        	\For{$j=1$ to $M$}
                	\State $C[i,j] = C[i,j] + \mu_j$;
                \EndFor
        \EndFor
        
        \State $\mathbf{Y} = (\mathbf{R}^\top \mathbf{C})^\top$;        
\end{algorithmic}
\normalfont
\end{algorithm}

\begin{algorithm}[t]
%\scriptsize
%\footnotesize
\small
\caption{Row vector reordering}
\label{alg:reordering}
%\algsetup{linenosize=\scriptsize}
\begin{algorithmic}[1]
	\Input  matrix $\mathbf{A}$
        \Output reordered matrix $\mathbf{A}^*$, column-wise mean vector $\mathbf{m}$
        
        \State $\mathbf{m} = \frac{1}{N} \sum_{i=1}^{N} \mathbf{A}_i$;
        
        \For{$i=1$ to $N$}
        	\State $\mathbf{A}_{i}^\prime=\mathbf{A}_{i} - \mathbf{m}$;
        \EndFor
        
        \State compute the projection vector $\mathbf{u}$ by Equation~(\ref{eq:CC-IPCA});
        
        \For{$i=1$ to $N$}
        	\State $s_i=\mathbf{A}_{i}^\prime \mathbf{u}^\top$;
        \EndFor
        
        \For{$i=1$ to $N$}
        	\State $s_i^\prime = \min(\lfloor m\frac{s_i-s_{\min}}{s_{\max}-s_{\min}}\rfloor,m-1)$;
        \EndFor
        
        \State perform bucket sort on the row vectors according to $s_i^\prime$;
        \State $\mathbf{R} = \mathbf{0}$; 
        
        \For{$i=1$ to $N$}
        	\State find the row vector $\mathbf{A}_j^\prime$ from the buckets where $j$ is its index;
                \State $R[j,i] = 1$;
        \EndFor
        
        \For{$i=1$ to $N$}
                \State $\mathbf{A}_i = \mathbf{A}_i^\prime$;
        \EndFor
        
        \State $\mathbf{A}^* = \mathbf{R} \mathbf{A}$;        
\end{algorithmic}
\normalfont
\end{algorithm}

\subsection{Algorithm}

Algorithm~\ref{alg:proposed} summarizes the proposed photonic matrix-multiplication procedure.
Unlike the original approach,
the proposed approach reformulates the matrix multiplication as
$\mathbf X^\top \mathbf W^\top=\mathbf Y^\top$.
It initializes $\mathbf A=\mathbf X^\top$, $\mathbf B=\mathbf W^\top$, and $\mathbf C=\mathbf0$ (line~1). 
It then reorders matrix $\mathbf{A}$ to obtain the reordered matrix $\mathbf{A}^*$ 
using Algorithm~\ref{alg:reordering} (line~3). 
Algorithm~\ref{alg:reordering} computes each mean-centered row vector $\mathbf{A}_i^\prime$ 
before reordering (lines~1--3).
It then computes the projection scores 
and performs bucket sort on the quantized scores (lines~4--9). 
It then constructs the reordered matrix $\mathbf{A}^*$ by forming the reordering
matrix $\mathbf R$ from the sorting result (lines~10--16).
After reordering, Algorithm~\ref{alg:proposed} 
partitions matrices $\mathbf{A}^*$ and $\mathbf{B}$ as in the original approach (lines~4--5).
For each block,
Algorithm~\ref{alg:proposed} computes and normalizes
the representative vector since the scaling method assumes a unit-norm representative vector (lines~6--8). 
Each row vector and the corresponding column vector are decomposed
into representative and residual components 
using the scaled representative vectors,
after which the residual vectors are quantized by ABFP (lines~9--16).
The photonic device computes the residual inner products, 
after which 
the representative component $\rho_{\mathbf a}\rho_{\mathbf b}$ is added back
according to Equation~(\ref{eq:scale_rep_ab_b_re}) (lines~16--20).
Since each row vector in matrix $\mathbf{A}^*$ is mean-centered, 
each element of the obtained matrix product corresponds to 
$\mathbf{A}_{i}^\prime \mathbf{B}_{j}$ in Equation~(\ref{eq:nomal_ab_inner}). 
Therefore, the corresponding mean offset $\mu_j$ is added to every output element (lines~21--25).
Finally, the original row order is recovered 
by applying $\mathbf R^\top$, 
and the output matrix $\mathbf Y$ is obtained by transposition (line~26).

The computational and memory costs of the proposed approach 
are summarized in the following lemma:
\begin{lemma}
\label{lem:pro_cost}
Let $\mathbf{A} \in \mathbb{R}^{N \times D}$ and $\mathbf{B} \in \mathbb{R}^{D \times M}$.
The proposed approach requires $O(ND + MD + NMD/n)$ time and
$O(ND + MD + NM)$ space.
\end{lemma}
\begin{proof}
Mean-centering the matrix $\mathbf{A}$ requires $O(ND)$ time.
The offset components are computed in $O(MD)$ time.
During reordering,
unit vector $\mathbf{u}$ and the projection scores of the row vectors 
can be computed in $O(ND)$ time.
The row vectors are reordered by bucket sort in $O(N)$ time.
For the scaling step,
the representative vector of each block matrix is computed in $O(n^2)$ time.
Since there are $ND/n^2$ block matrices, the total computational cost is $O(ND)$ 
for all representative vectors.
Similarly, all scaling parameters for vector $\mathbf{a}$ are computed in $O(ND)$ time.
Each scaling parameter $\rho_{\mathbf{b}}$ requires $O(n)$ time for a block vector.
Since there are $N/n$ blocks in each column in matrix $\mathbf{A}$ and 
$MD/n$ block vectors in matrix $\mathbf{B}$, 
all scaling parameters associated with $\mathbf{b}$
are computed in $O(NMD/n)$ time.
For each inner product, the proposed approach computes $\rho_{\mathbf{a}}\rho_{\mathbf{b}}$
on the digital processor in $O(1)$ time.
Since the number of inner products is $NMD/n$, their total computational cost is $O(NMD/n)$.
Therefore, the overall computational complexity is 
$O(ND + MD + NMD/n)$.
For the memory complexity,
centering matrix $\mathbf{A}$ requires $O(ND + MD)$ space.
During reordering, unit vector $\mathbf{u}$
and projection scores require $O(D)$ and $O(N)$ space, respectively.
Each representative vector consists of $n$ elements and requires $O(n)$ space.
Similarly, storing the scaling parameters
$\rho_{\mathbf{a}}$ and $\rho_{\mathbf{b}}$
requires $O(n)$ and $O(1)$ space, respectively.
The memory cost for matrices $\mathbf{A}$, $\mathbf{B}$, and $\mathbf{C}$
is $O(ND + MD + NM)$.
Therefore, the total memory complexity of the proposed approach is $O(ND + MD + NM)$.
\end{proof}

%The proof of this lemma is shown in the appendix.
Compared with the original photonic matrix-multiplication approach 
summarized in Lemma~\ref{lem:ori_cost}, 
the proposed approach 
improves inference accuracy without increasing either the computational or memory complexity.

\begin{table*}[t]
\centering
%\small
\footnotesize
%\scriptsize
\setlength{\tabcolsep}{2.5pt}
\caption{Inference accuracy [\%]. 
Cell colors indicate the absolute difference in inference accuracy 
from the FP32 baseline: green, $\leq 1$ percentage point; 
yellow, $>1$ and $\leq 5$ percentage points; orange, $>5$ and $\leq 10$ percentage points; 
red, $>10$ percentage points.
}
\vspace{-0.5em}
\label{tab:accuracy_performance}
\begin{tabular}{lr@{\hspace{2mm}} ccccc c @{\hspace{1.0mm}} ccccc c}
\toprule
\multirow{2}{*}{Model} & \multirow{2}{*}{$n$} & \multicolumn{5}{c}{Original} & & \multicolumn{5}{c}{Proposed} & \multirow{2}{*}{FP32} \\
\cmidrule(lr){3-7}\cmidrule(lr){9-13}
 & & $G=1$ & $G=2$ & $G=4$ & $G=8$ & $G=16$ & & $G=1$ & $G=2$ & $G=4$ & $G=8$ & $G=16$ & \\
\midrule
\multirow{3}{*}{ResNet50} & $8$ & \cellcolor{cYellow}75.73\stdc{0.46} & \cellcolor{cYellow}75.87\stdc{0.31} & \cellcolor{cOrange}72.00\stdc{0.69} & \cellcolor{cRed}1.20\stdc{0.00} & \cellcolor{cRed}0.20\stdc{0.00} & & \cellcolor{cGreen}77.27\stdc{0.46} & \cellcolor{cGreen}77.00\stdc{0.35} & \cellcolor{cYellow}76.07\stdc{0.12} & \cellcolor{cRed}8.73\stdc{3.95} & \cellcolor{cRed}0.53\stdc{0.46} & \multirow{3}{*}{77.20} \\
 & $32$ & \cellcolor{cRed}52.73\stdc{0.46} & \cellcolor{cOrange}72.13\stdc{0.90} & \cellcolor{cYellow}75.00\stdc{0.69} & \cellcolor{cOrange}69.13\stdc{0.76} & \cellcolor{cRed}26.73\stdc{0.12} & & \cellcolor{cGreen}76.67\stdc{0.61} & \cellcolor{cGreen}77.13\stdc{0.46} & \cellcolor{cGreen}77.13\stdc{0.46} & \cellcolor{cGreen}77.53\stdc{0.50} & \cellcolor{cYellow}74.80\stdc{1.40} &  \\
 & $128$ & \cellcolor{cRed}0.00\stdc{0.00} & \cellcolor{cRed}0.33\stdc{0.31} & \cellcolor{cRed}55.33\stdc{0.50} & \cellcolor{cYellow}72.73\stdc{0.12} & \cellcolor{cYellow}73.80\stdc{0.20} & & \cellcolor{cRed}0.27\stdc{0.46} & \cellcolor{cYellow}75.53\stdc{1.42} & \cellcolor{cGreen}77.00\stdc{0.00} & \cellcolor{cGreen}77.13\stdc{0.31} & \cellcolor{cGreen}77.20\stdc{0.00} &  \\
\midrule
\multirow{3}{*}{SSD-ResNet34} & $8$ & \cellcolor{cGreen}19.91\stdc{0.07} & \cellcolor{cYellow}17.14\stdc{0.02} & \cellcolor{cRed}5.85\stdc{0.01} & \cellcolor{cRed}0.00\stdc{0.00} & \cellcolor{cRed}0.00\stdc{0.00} & & \cellcolor{cGreen}19.95\stdc{0.01} & \cellcolor{cGreen}19.92\stdc{0.02} & \cellcolor{cYellow}17.80\stdc{0.02} & \cellcolor{cRed}0.45\stdc{0.01} & \cellcolor{cRed}0.00\stdc{0.00} & \multirow{3}{*}{20.05} \\
 & $32$ & \cellcolor{cGreen}19.31\stdc{0.02} & \cellcolor{cGreen}19.70\stdc{0.05} & \cellcolor{cYellow}17.12\stdc{0.01} & \cellcolor{cOrange}11.40\stdc{0.01} & \cellcolor{cRed}0.38\stdc{0.01} & & \cellcolor{cGreen}19.81\stdc{0.05} & \cellcolor{cGreen}19.88\stdc{0.02} & \cellcolor{cGreen}19.77\stdc{0.02} & \cellcolor{cYellow}18.00\stdc{0.04} & \cellcolor{cRed}3.95\stdc{0.01} &  \\
 & $128$ & \cellcolor{cRed}4.48\stdc{0.02} & \cellcolor{cYellow}16.85\stdc{0.01} & \cellcolor{cGreen}19.11\stdc{0.06} & \cellcolor{cGreen}19.50\stdc{0.02} & \cellcolor{cYellow}16.24\stdc{0.04} & & \cellcolor{cOrange}14.15\stdc{0.08} & \cellcolor{cYellow}18.57\stdc{0.07} & \cellcolor{cGreen}19.59\stdc{0.04} & \cellcolor{cGreen}19.69\stdc{0.03} & \cellcolor{cYellow}18.20\stdc{0.05} &  \\
\midrule

\multirow{3}{*}{3D U-Net} & $8$ & \cellcolor{cYellow}81.35\stdc{0.01} & \cellcolor{cYellow}80.58\stdc{0.01} & \cellcolor{cRed}64.56\stdc{0.01} & \cellcolor{cRed}49.28\stdc{0.00} & \cellcolor{cRed}45.63\stdc{0.00} & & \cellcolor{cYellow}81.28\stdc{0.01} & \cellcolor{cYellow}81.28\stdc{0.00} & \cellcolor{cYellow}80.90\stdc{0.02} & \cellcolor{cRed}70.66\stdc{0.32} & \cellcolor{cRed}47.17\stdc{0.22} & \multirow{3}{*}{85.36} \\
 & $32$ & \cellcolor{cYellow}81.64\stdc{0.01} & \cellcolor{cYellow}81.42\stdc{0.01} & \cellcolor{cOrange}79.54\stdc{0.01} & \cellcolor{cRed}66.01\stdc{0.01} & \cellcolor{cRed}44.56\stdc{0.01} & & \cellcolor{cYellow}81.24\stdc{0.02} & \cellcolor{cYellow}81.26\stdc{0.01} & \cellcolor{cYellow}81.22\stdc{0.01} & \cellcolor{cOrange}79.04\stdc{0.01} & \cellcolor{cRed}60.11\stdc{0.01} &  \\
 & $128$ & \cellcolor{cYellow}80.73\stdc{0.28} & \cellcolor{cYellow}81.43\stdc{0.03} & \cellcolor{cYellow}81.51\stdc{0.01} & \cellcolor{cYellow}81.06\stdc{0.00} & \cellcolor{cOrange}76.38\stdc{0.01} & & \cellcolor{cOrange}80.08\stdc{0.03} & \cellcolor{cYellow}80.97\stdc{0.02} & \cellcolor{cYellow}81.18\stdc{0.01} & \cellcolor{cYellow}81.18\stdc{0.01} & \cellcolor{cOrange}80.27\stdc{0.02} &  \\
\midrule
\multirow{3}{*}{RNN-T} & $8$ & \cellcolor{cGreen}96.33\stdc{0.02} & \cellcolor{cGreen}96.33\stdc{0.01} & \cellcolor{cYellow}94.98\stdc{0.01} & \cellcolor{cRed}22.10\stdc{0.05} & \cellcolor{cRed}0.00\stdc{0.00} & & \cellcolor{cGreen}96.28\stdc{0.02} & \cellcolor{cGreen}96.27\stdc{0.02} & \cellcolor{cGreen}96.20\stdc{0.02} & \cellcolor{cOrange}90.74\stdc{0.05} & \cellcolor{cRed}2.09\stdc{0.03} & \multirow{3}{*}{96.34} \\
 & $32$ & \cellcolor{cGreen}95.82\stdc{0.06} & \cellcolor{cGreen}96.18\stdc{0.10} & \cellcolor{cGreen}96.24\stdc{0.02} & \cellcolor{cGreen}96.03\stdc{0.01} & \cellcolor{cYellow}93.43\stdc{0.04} & & \cellcolor{cGreen}96.25\stdc{0.04} & \cellcolor{cGreen}96.30\stdc{0.03} & \cellcolor{cGreen}96.29\stdc{0.01} & \cellcolor{cGreen}96.10\stdc{0.02} & \cellcolor{cYellow}94.88\stdc{0.07} &  \\
 & $128$ & \cellcolor{cRed}0.00\stdc{0.00} & \cellcolor{cYellow}92.82\stdc{0.30} & \cellcolor{cGreen}95.88\stdc{0.04} & \cellcolor{cGreen}96.19\stdc{0.08} & \cellcolor{cGreen}96.23\stdc{0.02} & & \cellcolor{cOrange}90.65\stdc{0.42} & \cellcolor{cGreen}95.82\stdc{0.09} & \cellcolor{cGreen}96.22\stdc{0.01} & \cellcolor{cGreen}96.29\stdc{0.02} & \cellcolor{cGreen}96.28\stdc{0.04} &  \\
\midrule
\multirow{3}{*}{BERT-Large} & $8$ & \cellcolor{cGreen}92.85\stdc{0.29} & \cellcolor{cGreen}92.88\stdc{0.32} & \cellcolor{cGreen}92.65\stdc{0.53} & \cellcolor{cRed}9.25\stdc{0.15} & \cellcolor{cRed}6.66\stdc{0.02} & & \cellcolor{cGreen}92.86\stdc{0.35} & \cellcolor{cGreen}92.88\stdc{0.35} & \cellcolor{cGreen}92.79\stdc{0.43} & \cellcolor{cRed}14.19\stdc{17.87} & \cellcolor{cRed}0.40\stdc{0.11} & \multirow{3}{*}{93.16} \\
 & $32$ & \cellcolor{cYellow}89.61\stdc{0.22} & \cellcolor{cGreen}92.26\stdc{0.16} & \cellcolor{cGreen}92.47\stdc{0.05} & \cellcolor{cGreen}92.39\stdc{0.43} & \cellcolor{cRed}8.34\stdc{5.84} & & \cellcolor{cGreen}92.43\stdc{0.34} & \cellcolor{cGreen}92.82\stdc{0.40} & \cellcolor{cGreen}92.82\stdc{0.38} & \cellcolor{cGreen}92.54\stdc{0.39} & \cellcolor{cRed}22.56\stdc{29.01} &  \\
 & $128$ & \cellcolor{cRed}6.55\stdc{0.15} & \cellcolor{cRed}3.84\stdc{0.08} & \cellcolor{cOrange}87.03\stdc{0.17} & \cellcolor{cYellow}91.82\stdc{0.16} & \cellcolor{cGreen}92.49\stdc{0.19} &  &\cellcolor{cRed}5.96\stdc{2.87} & \cellcolor{cYellow}89.27\stdc{1.13} & \cellcolor{cGreen}92.26\stdc{0.46} & \cellcolor{cGreen}92.59\stdc{0.39} & \cellcolor{cYellow}91.74\stdc{1.09} &  \\
\midrule
\multirow{3}{*}{DLRM} & $8$ & \cellcolor{cGreen}80.84\stdc{0.02} & \cellcolor{cGreen}80.87\stdc{0.01} & \cellcolor{cGreen}80.86\stdc{0.01} & \cellcolor{cYellow}78.76\stdc{0.01} & \cellcolor{cOrange}72.67\stdc{0.00} & & \cellcolor{cGreen}80.91\stdc{0.01} & \cellcolor{cGreen}80.91\stdc{0.00} & \cellcolor{cGreen}80.90\stdc{0.00} & \cellcolor{cGreen}80.08\stdc{0.03} & \cellcolor{cYellow}76.63\stdc{0.07} & \multirow{3}{*}{80.93} \\
 & $32$ & \cellcolor{cGreen}79.93\stdc{0.02} & \cellcolor{cGreen}80.59\stdc{0.03} & \cellcolor{cGreen}80.78\stdc{0.01} & \cellcolor{cGreen}80.83\stdc{0.01} & \cellcolor{cGreen}80.46\stdc{0.00} & & \cellcolor{cGreen}80.82\stdc{0.03} & \cellcolor{cGreen}80.89\stdc{0.00} & \cellcolor{cGreen}80.91\stdc{0.01} & \cellcolor{cGreen}80.91\stdc{0.01} & \cellcolor{cGreen}80.84\stdc{0.01} &  \\
 & $128$ & \cellcolor{cOrange}71.65\stdc{0.12} & \cellcolor{cYellow}77.86\stdc{0.07} & \cellcolor{cYellow}79.89\stdc{0.06} & \cellcolor{cGreen}80.57\stdc{0.01} & \cellcolor{cGreen}80.71\stdc{0.01} & & \cellcolor{cYellow}79.56\stdc{0.04} & \cellcolor{cGreen}80.53\stdc{0.01} & \cellcolor{cGreen}80.81\stdc{0.01} & \cellcolor{cGreen}80.88\stdc{0.01} & \cellcolor{cGreen}80.90\stdc{0.01} &  \\
\bottomrule
\end{tabular}
\\
\vspace{1.0em}
(1) $b_\mathbf{A}=6$, $b_\mathbf{B}=6$, and $b_\mathbf{C}=8$
\vspace{1.0em}

\begin{tabular}{lr@{\hspace{2mm}} ccccc c @{\hspace{1.0mm}} ccccc c}
\toprule
\multirow{2}{*}{Model} & \multirow{2}{*}{$n$} & \multicolumn{5}{c}{Original} & & \multicolumn{5}{c}{Proposed} & \multirow{2}{*}{FP32} \\
\cmidrule(lr){3-7}\cmidrule(lr){9-13}
 & & $G=1$ & $G=2$ & $G=4$ & $G=8$ & $G=16$ & & $G=1$ & $G=2$ & $G=4$ & $G=8$ & $G=16$ & \\
\midrule
\multirow{3}{*}{ResNet50} & $8$ & \cellcolor{cGreen}76.40\stdc{0.53} & \cellcolor{cGreen}76.27\stdc{0.42} & \cellcolor{cYellow}73.93\stdc{0.12} & \cellcolor{cRed}1.00\stdc{0.00} & \cellcolor{cRed}0.20\stdc{0.00} & & \cellcolor{cGreen}77.07\stdc{0.12} & \cellcolor{cGreen}77.00\stdc{0.20} & \cellcolor{cYellow}75.93\stdc{0.23} & \cellcolor{cRed}8.93\stdc{3.93} & \cellcolor{cRed}0.40\stdc{0.40} & \multirow{3}{*}{77.20} \\
 & $32$ & \cellcolor{cRed}55.40\stdc{1.06} & \cellcolor{cYellow}73.40\stdc{0.40} & \cellcolor{cGreen}76.20\stdc{0.20} & \cellcolor{cOrange}70.47\stdc{0.64} & \cellcolor{cRed}35.60\stdc{0.20} & & \cellcolor{cGreen}77.13\stdc{0.64} & \cellcolor{cGreen}76.73\stdc{0.90} & \cellcolor{cGreen}76.47\stdc{0.23} & \cellcolor{cGreen}77.13\stdc{0.70} & \cellcolor{cYellow}74.33\stdc{1.50} &  \\
 & $128$ & \cellcolor{cRed}0.00\stdc{0.00} & \cellcolor{cRed}0.27\stdc{0.31} & \cellcolor{cRed}57.07\stdc{0.81} & \cellcolor{cYellow}73.20\stdc{0.69} & \cellcolor{cYellow}74.13\stdc{0.46} & & \cellcolor{cRed}0.40\stdc{0.20} & \cellcolor{cYellow}74.73\stdc{0.50} & \cellcolor{cGreen}77.47\stdc{0.70} & \cellcolor{cGreen}77.67\stdc{0.61} & \cellcolor{cGreen}77.27\stdc{0.23} &  \\
\midrule
\multirow{3}{*}{SSD-ResNet34} & $8$ & \cellcolor{cGreen}19.98\stdc{0.03} & \cellcolor{cYellow}17.14\stdc{0.01} & \cellcolor{cRed}5.77\stdc{0.01} & \cellcolor{cRed}0.00\stdc{0.00} & \cellcolor{cRed}0.00\stdc{0.00} & & \cellcolor{cGreen}19.95\stdc{0.04} & \cellcolor{cGreen}19.91\stdc{0.03} & \cellcolor{cYellow}17.79\stdc{0.03} & \cellcolor{cRed}0.45\stdc{0.02} & \cellcolor{cRed}0.00\stdc{0.00} & \multirow{3}{*}{20.05} \\
 & $32$ & \cellcolor{cGreen}19.39\stdc{0.02} & \cellcolor{cGreen}19.78\stdc{0.02} & \cellcolor{cYellow}17.16\stdc{0.03} & \cellcolor{cOrange}11.49\stdc{0.01} & \cellcolor{cRed}0.36\stdc{0.01} & & \cellcolor{cGreen}19.76\stdc{0.04} & \cellcolor{cGreen}19.93\stdc{0.06} & \cellcolor{cGreen}19.80\stdc{0.03} & \cellcolor{cYellow}17.99\stdc{0.02} & \cellcolor{cRed}3.97\stdc{0.03} &  \\
 & $128$ & \cellcolor{cRed}4.64\stdc{0.05} & \cellcolor{cYellow}17.03\stdc{0.06} & \cellcolor{cYellow}19.33\stdc{0.01} & \cellcolor{cGreen}19.73\stdc{0.05} & \cellcolor{cYellow}16.55\stdc{0.01} & & \cellcolor{cOrange}14.21\stdc{0.14} & \cellcolor{cYellow}18.61\stdc{0.07} & \cellcolor{cGreen}19.63\stdc{0.06} & \cellcolor{cGreen}19.74\stdc{0.06} & \cellcolor{cYellow}18.26\stdc{0.01} &  \\
\midrule
\multirow{3}{*}{3D U-Net} & $8$ & \cellcolor{cYellow}81.39\stdc{0.00} & \cellcolor{cYellow}80.54\stdc{0.00} & \cellcolor{cRed}64.42\stdc{0.01} & \cellcolor{cRed}49.23\stdc{0.01} & \cellcolor{cRed}45.56\stdc{0.00} & & \cellcolor{cYellow}81.28\stdc{0.00} & \cellcolor{cYellow}81.28\stdc{0.00} & \cellcolor{cYellow}80.90\stdc{0.01} & \cellcolor{cRed}70.69\stdc{0.26} & \cellcolor{cRed}47.32\stdc{0.04} & \multirow{3}{*}{85.36} \\
 & $32$ & \cellcolor{cYellow}81.57\stdc{0.01} & \cellcolor{cYellow}81.44\stdc{0.01} & \cellcolor{cOrange}79.45\stdc{0.01} & \cellcolor{cRed}65.67\stdc{0.00} & \cellcolor{cRed}44.68\stdc{0.01} & & \cellcolor{cYellow}81.23\stdc{0.01} & \cellcolor{cYellow}81.27\stdc{0.01} & \cellcolor{cYellow}81.22\stdc{0.01} & \cellcolor{cOrange}78.86\stdc{0.28} & \cellcolor{cRed}60.14\stdc{0.03} &  \\
 & $128$ & \cellcolor{cYellow}80.42\stdc{0.29} & \cellcolor{cYellow}81.35\stdc{0.02} & \cellcolor{cYellow}81.46\stdc{0.01} & \cellcolor{cYellow}81.08\stdc{0.00} & \cellcolor{cOrange}76.44\stdc{0.01} & & \cellcolor{cOrange}80.15\stdc{0.03} & \cellcolor{cYellow}81.03\stdc{0.04} & \cellcolor{cYellow}81.24\stdc{0.01} & \cellcolor{cYellow}81.23\stdc{0.01} & \cellcolor{cOrange}80.29\stdc{0.01} &  \\
\midrule
\multirow{3}{*}{RNN-T} & $8$ & \cellcolor{cGreen}96.34\stdc{0.01} & \cellcolor{cGreen}96.33\stdc{0.01} & \cellcolor{cYellow}94.97\stdc{0.01} & \cellcolor{cRed}21.69\stdc{0.10} & \cellcolor{cRed}0.00\stdc{0.00} & & \cellcolor{cGreen}96.26\stdc{0.02} & \cellcolor{cGreen}96.25\stdc{0.01} & \cellcolor{cGreen}96.23\stdc{0.01} & \cellcolor{cOrange}90.77\stdc{0.06} & \cellcolor{cRed}2.08\stdc{0.02} & \multirow{3}{*}{96.34} \\
 & $32$ & \cellcolor{cGreen}95.75\stdc{0.03} & \cellcolor{cGreen}96.22\stdc{0.05} & \cellcolor{cGreen}96.30\stdc{0.03} & \cellcolor{cGreen}96.04\stdc{0.03} & \cellcolor{cYellow}93.43\stdc{0.02} & & \cellcolor{cGreen}96.26\stdc{0.06} & \cellcolor{cGreen}96.30\stdc{0.03} & \cellcolor{cGreen}96.30\stdc{0.01} & \cellcolor{cGreen}96.09\stdc{0.02} & \cellcolor{cGreen}94.89\stdc{0.05} &  \\
 & $128$ & \cellcolor{cRed}0.00\stdc{0.00} & \cellcolor{cYellow}92.80\stdc{0.21} & \cellcolor{cGreen}95.92\stdc{0.07} & \cellcolor{cGreen}96.26\stdc{0.01} & \cellcolor{cGreen}96.29\stdc{0.03} & & \cellcolor{cOrange}90.84\stdc{0.44} & \cellcolor{cGreen}95.82\stdc{0.11} & \cellcolor{cGreen}96.20\stdc{0.04} & \cellcolor{cGreen}96.30\stdc{0.02} & \cellcolor{cGreen}96.28\stdc{0.02} &  \\
\midrule
\multirow{3}{*}{BERT-Large} & $8$ & \cellcolor{cGreen}92.76\stdc{0.35} & \cellcolor{cGreen}92.84\stdc{0.30} & \cellcolor{cGreen}92.59\stdc{0.43} & \cellcolor{cRed}9.20\stdc{0.15} & \cellcolor{cRed}6.74\stdc{0.04} & & \cellcolor{cGreen}92.81\stdc{0.35} & \cellcolor{cGreen}92.84\stdc{0.35} & \cellcolor{cGreen}92.80\stdc{0.43} & \cellcolor{cRed}14.10\stdc{17.98} & \cellcolor{cRed}0.37\stdc{0.12} & \multirow{3}{*}{93.16} \\
 & $32$ & \cellcolor{cYellow}89.74\stdc{0.10} & \cellcolor{cGreen}92.32\stdc{0.09} & \cellcolor{cGreen}92.56\stdc{0.04} & \cellcolor{cGreen}92.57\stdc{0.32} & \cellcolor{cRed}7.75\stdc{5.91} & & \cellcolor{cGreen}92.42\stdc{0.30} & \cellcolor{cGreen}92.82\stdc{0.34} & \cellcolor{cGreen}92.81\stdc{0.32} & \cellcolor{cGreen}92.56\stdc{0.45} & \cellcolor{cRed}22.74\stdc{29.54} &  \\
 & $128$ & \cellcolor{cRed}6.43\stdc{0.36} & \cellcolor{cRed}4.28\stdc{0.11} & \cellcolor{cYellow}87.47\stdc{0.20} & \cellcolor{cYellow}92.03\stdc{0.16} & \cellcolor{cGreen}92.51\stdc{0.12} & & \cellcolor{cRed}4.59\stdc{2.07} & \cellcolor{cYellow}89.20\stdc{1.24} & \cellcolor{cGreen}92.31\stdc{0.36} & \cellcolor{cGreen}92.71\stdc{0.37} & \cellcolor{cYellow}91.93\stdc{1.00} &  \\
\midrule
\multirow{3}{*}{DLRM} & $8$ & \cellcolor{cGreen}80.87\stdc{0.02} & \cellcolor{cGreen}80.90\stdc{0.01} & \cellcolor{cGreen}80.89\stdc{0.00} & \cellcolor{cYellow}78.82\stdc{0.00} & \cellcolor{cOrange}72.68\stdc{0.00} & & \cellcolor{cGreen}80.92\stdc{0.01} & \cellcolor{cGreen}80.93\stdc{0.00} & \cellcolor{cGreen}80.91\stdc{0.00} & \cellcolor{cGreen}80.05\stdc{0.02} & \cellcolor{cYellow}76.62\stdc{0.04} & \multirow{3}{*}{80.93} \\
 & $32$ & \cellcolor{cGreen}79.98\stdc{0.02} & \cellcolor{cGreen}80.67\stdc{0.02} & \cellcolor{cGreen}80.87\stdc{0.01} & \cellcolor{cGreen}80.91\stdc{0.01} & \cellcolor{cGreen}80.56\stdc{0.01} & & \cellcolor{cGreen}80.82\stdc{0.01} & \cellcolor{cGreen}80.91\stdc{0.01} & \cellcolor{cGreen}80.92\stdc{0.01} & \cellcolor{cGreen}80.92\stdc{0.00} & \cellcolor{cGreen}80.85\stdc{0.00} &  \\
 & $128$ & \cellcolor{cOrange}71.65\stdc{0.06} & \cellcolor{cYellow}78.00\stdc{0.15} & \cellcolor{cGreen}79.97\stdc{0.08} & \cellcolor{cGreen}80.67\stdc{0.00} & \cellcolor{cGreen}80.81\stdc{0.01} & & \cellcolor{cYellow}79.55\stdc{0.03} & \cellcolor{cGreen}80.57\stdc{0.06} & \cellcolor{cGreen}80.83\stdc{0.01} & \cellcolor{cGreen}80.90\stdc{0.01} & \cellcolor{cGreen}80.92\stdc{0.00} &  \\
\bottomrule
\end{tabular}
\\
\vspace{1.0em}
(2) $b_\mathbf{A}=8$, $b_\mathbf{B}=8$
\end{table*}

\section{Experimental Evaluation}
\label{sec:experimental}

We conducted experiments to evaluate the effectiveness
of the proposed approach in photonic matrix multiplication.
Throughout this section,
``Original'' refers to the original photonic matrix-multiplication approach,
whereas ``Proposed'' denotes our approach.
Following the experimental setup in the previous study
\cite{DBLP:journals/corr/abs-2205-06287},
we evaluated each approach on the following six representative DNN models 
covering a broad range of AI applications,
including image classification, object detection, image segmentation, speech recognition, 
question answering, and recommendation:

\begin{itemize}
\item \textbf{ResNet50} is a convolutional neural network for large-scale image classification. 
      We evaluated it on the ImageNet dataset, which contains images from $1{,}000$ 
      object categories and is a standard benchmark for image classification.
\item \textbf{SSD-ResNet34} is a single-stage object detection model 
      that combines the Single Shot Detector framework with a ResNet34 backbone. 
      We evaluated it on the MS COCO dataset, a widely used benchmark for 
      object detection containing diverse scenes with bounding-box annotations.
\item \textbf{3D U-Net} is a volumetric convolutional neural network designed 
      for medical image segmentation. 
      We evaluated it on the BRaTS 2019 dataset, which consists of multimodal 
      brain MRI scans with tumor segmentation annotations.
\item \textbf{RNN-T} is an end-to-end automatic speech recognition model. 
      We evaluated it on the LibriSpeech dataset, 
      a standard benchmark containing approximately $1{,}000$ hours of English read speech.
\item \textbf{BERT-Large} is a Transformer-based language model widely used for 
      natural language understanding tasks. 
      We evaluated it on the SQuAD v1.1 dataset, a machine reading comprehension benchmark 
      in which the model predicts answer spans from Wikipedia passages.
\item \textbf{DLRM} is designed for large-scale recommendation systems 
      by modeling interactions between dense and sparse features. 
      We evaluated it on the Criteo Terabyte (1TB) Click Logs dataset, 
      a standard benchmark for click-through rate prediction.
\end{itemize}

To isolate the effects of each approach, 
we evaluated all models without fine-tuning, 
following prior studies \cite{DBLP:journals/corr/abs-2205-06287,DBLP:journals/corr/abs-2307-03712}. 

The standard evaluation methodology used in previous studies is 
to perform hardware-aware simulations \cite{DBLP:conf/nips/GuZFJCP21,DBLP:conf/icton/DuzgolKPR23,DBLP:journals/jetc/DemirkiranEWEMHBRJB23}. 
This is because photonic accelerators have not yet been widely deployed commercially,
primarily due to 
physical challenges such as thermal drift and fabrication variations, 
as well as the hardware overhead of high-precision DACs and ADCs \cite{Huang31122022,moss_2025,11284869}.
The hardware-aware simulation models DAC/ADC quantization errors
and analog photonic noise.
This simulation framework enables a fair comparison of each approach
under identical photonic hardware settings.
Following prior studies \cite{DBLP:conf/dac/RekhiZNLVWKDG19,DBLP:journals/corr/abs-2205-06287},
we modeled analog photonic noise as an additive, 
zero-mean Gaussian random variable independent of the matrix values, 
with variance $\sigma^2=(n\delta_{\mathbf{C}})^2/12$.
Following prior work \cite{DBLP:journals/corr/abs-2205-06287},
we report inference accuracy and computation error in the experiments. 
This is because accuracy degradation is the critical bottleneck in
realizing photonic accelerators \cite{10609496,10.1117/1.AP.8.1.016005}, 
and hardware-dependent metrics such as inference latency, 
throughput, and energy consumption cannot be directly measured 
without physical photonic hardware, 
which is not yet widely available commercially \cite{DBLP:conf/ispass/AndrulisCSES24}.
We implemented the methods in C++ and Python and conducted all experiments 
on a Linux server equipped with a single NVIDIA GH200 GPU and 96 GB of memory.
We repeated each experiment five times and report the mean and standard deviation of the results.

\subsection{Inference Accuracy}

This experiment evaluated the inference accuracy of each approach.
Table~\ref{tab:accuracy_performance} reports the results for gain factors
$G=\{1,2,4,8,16\}$ and block lengths $n=\{8,32,128\}$.
The quantization bit widths were fixed at
$b_{\mathbf{A}}=6$, $b_{\mathbf{B}}=6$, and $b_{\mathbf{C}}=8$ in 
Table~\ref{tab:accuracy_performance}-1 and 
$b_{\mathbf{A}}=8$, $b_{\mathbf{B}}=8$, and $b_{\mathbf{C}}=8$ in 
Table~\ref{tab:accuracy_performance}-2 
by following the previous paper \cite{DBLP:journals/corr/abs-2205-06287}. 
In this table, 
``FP32'' denotes the inference accuracy without quantization or analog photonic noise.
Results with higher quantization bit widths are provided in the appendix.

\begin{table*}[t]
\centering
\footnotesize
%\scriptsize
\setlength{\tabcolsep}{9.0pt}
\caption{Inference accuracy under different analog photonic noise levels [\%].
The higher accuracy is shown in bold.}
\vspace{-0.5em}
\label{tab:noise_performance}
\begin{tabular}{clcccccc}
\toprule
Noise variance & Approach & ResNet50 & $\!\!$SSD-ResNet34$\!\!$ & 3D U-Net & RNN-T & BERT-Large & DLRM\\
\midrule
\multirow{2}{*}{$(\frac{1}{4}\sigma)^2$} & Original & 75.73\std{0.81} & 17.16\std{0.02} & 79.36\std{0.01} & 96.29\std{0.04} & 92.56\std{0.09} & 80.80\std{0.00} \\
 & Proposed & \textbf{76.40\std{0.20}} & \textbf{19.77\std{0.02}} & \textbf{81.21\std{0.01}} & \textbf{96.30\std{0.04}} & \textbf{92.65\std{0.09}} & \textbf{80.91\std{0.00}}\\
 \midrule
\multirow{2}{*}{$(\frac{1}{2}\sigma)^2$} & Original & 76.07\std{0.31} & 17.15\std{0.02} & 79.42\std{0.00} & 96.28\std{0.02} & 92.49\std{0.03} & 80.80\std{0.00} \\
 & Proposed & \textbf{76.27\std{0.31}} & \textbf{19.76\std{0.03}} & \textbf{81.22\std{0.00}} & \textbf{96.30\std{0.01}} & \textbf{92.66\std{0.04}} & \textbf{80.91\std{0.00}}\\
 \midrule
\multirow{2}{*}{$\sigma^2$} & Original & 75.00\std{0.69} & 17.12\std{0.01} & 79.54\std{0.01} & 96.24\std{0.02} & 92.47\std{0.05} & 80.78\std{0.01} \\
 & Proposed & \textbf{77.13\std{0.46}} & \textbf{19.77\std{0.02}} & \textbf{81.22\std{0.01}} & \textbf{96.29\std{0.01}} & \textbf{92.82\std{0.38}} & \textbf{80.91\std{0.01}}\\
 \midrule
\multirow{2}{*}{$(2\sigma)^2$} & Original & 65.87\std{0.81} & 16.76\std{0.04} & 79.85\std{0.01} & 95.93\std{0.07} & 91.08\std{0.10} & 80.35\std{0.04} \\
 & Proposed & \textbf{77.13\std{0.23}} & \textbf{19.70\std{0.03}} & \textbf{81.18\std{0.01}} & \textbf{96.25\std{0.01}} & \textbf{92.49\std{0.18}} & \textbf{80.86\std{0.01}}\\
 \midrule
\multirow{2}{*}{$(4\sigma)^2$} & Original & 0.13\std{0.12} & 9.34\std{0.07} & 80.55\std{0.03} & 24.13\std{13.46} & 2.42\std{0.26} & 75.38\std{0.05} \\
 & Proposed & \textbf{68.67\std{0.64}} & \textbf{18.10\std{0.06}} & \textbf{80.87\std{0.03}} & \textbf{94.19\std{0.14}} & \textbf{71.90\std{1.33}} & \textbf{80.24\std{0.02}}\\
\bottomrule
\end{tabular}
\vspace{-0.0em}
\end{table*}

Table~\ref{tab:accuracy_performance} shows that the proposed approach generally 
achieves higher inference accuracy than the original approach across 
the evaluated DNNs and operating configurations. 
The proposed approach computes the representative-vector component digitally 
at high precision and processes the residual component using photonic hardware, 
thereby reducing the effect of quantization errors 
and analog photonic noise on the residual computation.
An important observation is that 
the proposed approach achieves high inference accuracy even with $G=1$,
which requires no analog gain amplification.
For example, for ResNet50 with $n=32$ in Table~\ref{tab:accuracy_performance}-1, 
it improves inference accuracy from $52.73\%$ to $76.67\%$ at $G=1$.
Given that larger gain factors may increase power consumption \cite{8110629}, 
this result suggests that the proposed approach can reduce energy consumption
for analog gain amplification in some configurations.
In addition, 
the proposed approach maintains competitive accuracy across multiple gain 
factors and block lengths, 
offering additional flexibility in selecting operating parameters 
without modifying the photonic hardware.

The experimental results also suggest that the gain factor yielding the highest 
inference accuracy depends on the block length. 
For small block lengths, smaller gain factors often achieve higher accuracy, 
suggesting that large gains are unnecessary to use the ADC dynamic range effectively. 
For larger block lengths, each inner product sums more element-wise products, 
whose positive and negative contributions can partially cancel. 
This cancellation can reduce the magnitude of the resulting inner products. 
Increasing the gain factor can then improve the effective ADC resolution 
for these smaller values and reduce quantization errors. 
Overall, the results suggest that larger block lengths may favor larger gain factors, 
although the best setting varies across models.

Table~\ref{tab:accuracy_performance} shows that 
the proposed approach also maintains relatively high inference accuracy at a large block length ($n=128$) 
in several configurations where the original approach performs poorly. 
Prior work \cite{RePEc:nat:nature:v:640:y:2025:i:8058:d:10.1038_s41586-025-08854-x} 
reports that larger block lengths can improve photonic matrix-multiplication 
throughput by increasing hardware utilization. 
Therefore, these results suggest that the proposed approach may enable larger block lengths 
without severe accuracy degradation, potentially improving hardware utilization and throughput.

\subsection{Robustness to Analog Photonic Noise}

The proposed approach computes the representative-vector component 
to each inner product digitally at high precision, 
leaving the residual component to be evaluated by the photonic hardware.
To evaluate its robustness to analog photonic noise, we varied the noise variance 
in this experiment.
Table~\ref{tab:noise_performance} reports 
the inference accuracy at five noise-variance levels,
$\{ (\frac{1}{4}\sigma)^2, (\frac{1}{2}\sigma)^2, \sigma^2, (2\sigma)^2, (4\sigma)^2\}$, 
with the gain factor set to $G = 4$ and the block length set to $n = 32$.

Table~\ref{tab:noise_performance} shows that 
the proposed approach achieves higher inference accuracy than the original approach 
across all evaluated noise levels and models.
The difference is particularly pronounced at the highest noise level, $(4\sigma)^2$.
For example, 
for ResNet50, the original approach reduces inference accuracy to $0.13\%$, 
whereas the proposed approach still achieves $68.67\%$. 
This robustness is consistent with the proposed decomposition, 
in which the representative-vector component is computed digitally at high precision 
and the photonic hardware processes the residual component.
Because the residual vectors have smaller norms, 
errors in their photonic computation have a more limited effect on the final inner products.
These results indicate that the proposed approach improves robustness 
to analog photonic noise under the evaluated configurations.

\subsection{Ablation Study}

The proposed approach combines two techniques: 
row-vector reordering and representative-vector scaling.
Reordering groups similar activation vectors into the same block, 
improving the representative vectors' approximation of the original row vectors.
Scaling reduces the norms of the residual vectors processed by the photonic hardware.
Table~\ref{tab:ablation} 
presents an ablation study of these techniques, where
``w/o reordering'' and ``w/o scaling'' disable reordering and scaling, respectively.
We set the gain factor to $G = 4$ and the block length to $n = 32$.

As shown in Table~\ref{tab:ablation},
removing reordering reduces inference accuracy for most models. 
For 3D U-Net, disabling row-vector reordering slightly improves inference accuracy. 
One possible explanation is that, in 3D U-Net, the activation vectors exhibit heterogeneous spatial 
and semantic patterns that are not fully captured 
by their projections onto the first principal component. 
Consequently, ordering the vectors by their projection scores may not sufficiently 
increase within-block similarity or reduce the residual norms. 
On the other hand, removing scaling reduces inference accuracy for all six models, 
indicating that representative-vector scaling plays a consistent role 
to improve inference accuracy in the proposed approach.
By reducing residual norms, 
scaling can limit the effects of quantization errors 
and analog photonic noise on photonic computation.
With both techniques enabled, the proposed approach achieves 
the highest accuracy for five of the six models.
Overall, the two techniques offer complementary benefits: 
reordering can improve representative-vector quality, 
whereas scaling reduces the norms of the residual vectors processed 
by the photonic hardware.

\begin{table}[t]
\centering
%\small
\footnotesize
%\scriptsize
\setlength{\tabcolsep}{6pt}
\caption{Ablation study [\%]. Best results are shown in bold.}
\vspace{-0.5em}
\label{tab:ablation}
\begin{tabular}{lccc}
\toprule
Model & $\!$w/o reordering$\!$ & w/o scaling & Proposed \\
\midrule
ResNet50 & 74.73\std{0.12} & 67.73\std{1.01} & \textbf{77.13\std{0.46}} \\
SSD-ResNet34 & 17.04\std{0.02} & 17.67\std{0.01} & \textbf{19.77\std{0.02}} \\
3D U-Net & \textbf{81.54\std{0.01}} & 79.76\std{0.01} & 81.22\std{0.01} \\
RNN-T & 96.30\std{0.00} & 96.30\std{0.00} & \textbf{96.31\std{0.01}} \\
BERT-Large & 92.62\std{0.04} & 92.68\std{0.08} & \textbf{92.82\std{0.38}} \\
DLRM & 80.82\std{0.01} & 80.76\std{0.02} & \textbf{80.91\std{0.01}} \\
\bottomrule
\end{tabular}
\vspace{-1.0em}
\end{table}

\begin{figure*}[t]
\begin{center}
\small
\includegraphics[width=175mm,clip]{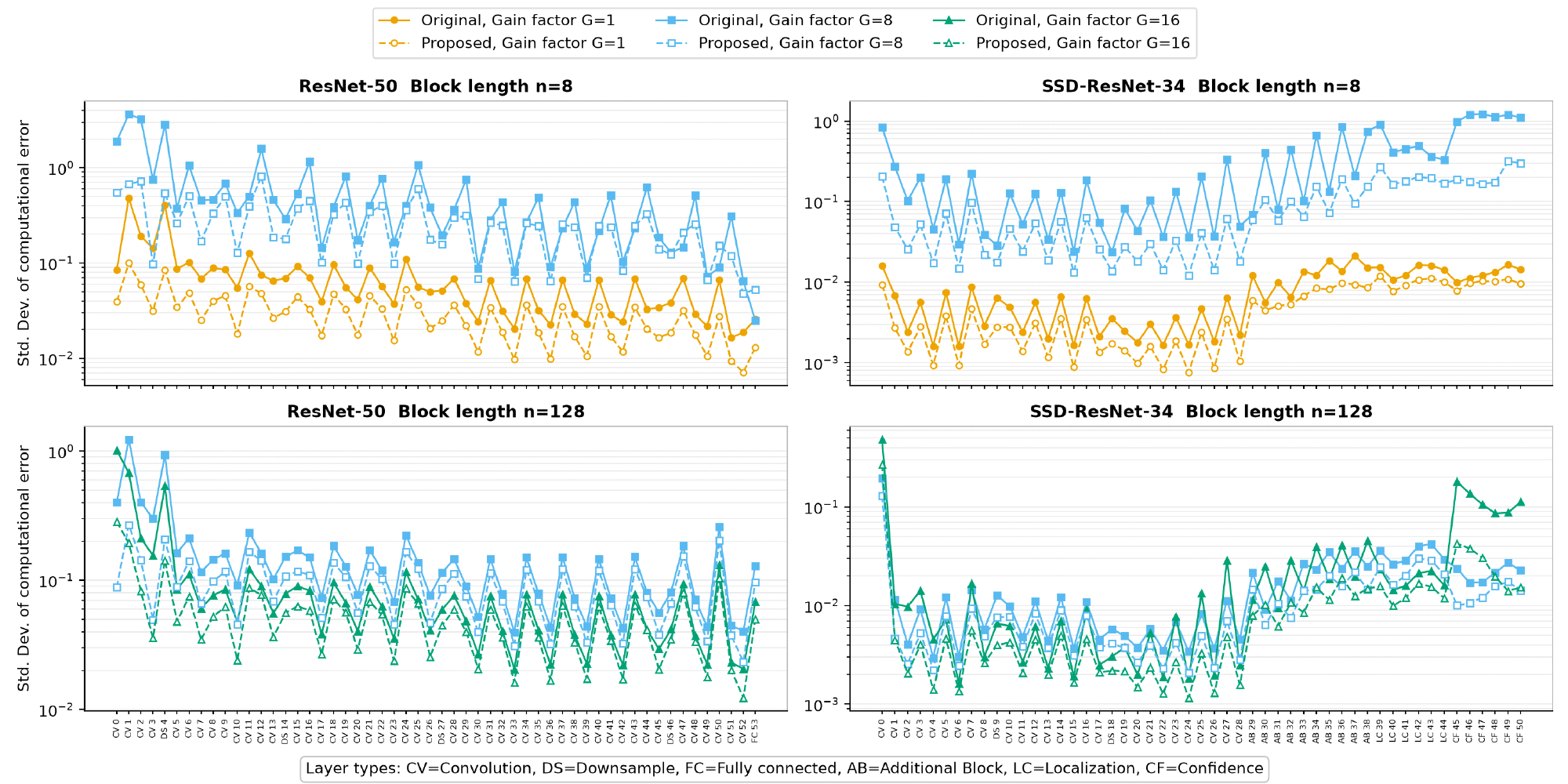}
\vspace{-1.0em}
\caption{Standard deviation of inner-product computation errors relative to FP32.}
\label{fig:comp_acc}
\normalsize
\end{center}
\vspace{-0.0em}
\end{figure*}

\subsection{Inner-product Computation Error}

We compared the inner-product computation errors of each approach.
Following the experimental protocol of the previous study \cite{DBLP:journals/corr/abs-2205-06287},
we show 
the standard deviation of the inner-product computation error with respect to the FP32 implementation 
for each layer of ResNet50 and SSD-ResNet34 in Figure~\ref{fig:comp_acc}.
We set $b_{\mathbf{A}}=8$, $b_{\mathbf{B}}=8$, and $b_{\mathbf{C}}=8$,
as in the previous study \cite{DBLP:journals/corr/abs-2205-06287}.
For $n=8$, we compared $G=\{1,8\}$, whereas for $n=128$, we compared $G=\{8,16\}$,
following the previous study \cite{DBLP:journals/corr/abs-2205-06287}.

For the small block length ($n=8$),
both the original and proposed approaches achieve lower computation errors 
with the smaller gain factor ($G=1$).
Since each vector is normalized before quantization,
the inner products computed by 
the photonic device lie within $[-n,n]$.
Therefore, when the block length is small, 
the range of possible inner-product values is narrow, 
so a relatively small gain is sufficient to effectively utilize the ADC dynamic range.
In contrast, for the large block length ($n=128$),
using the larger gain factor ($G=16$) 
provides comparable or lower computation errors than the smaller gain factor ($G=8$)
for both approaches.
As the block length increases,
the inner product is computed by summing many positive and negative terms
whose contributions tend to cancel each other,
resulting in smaller-magnitude inner products.
Consequently, for larger block lengths,
the larger gain improves the effective ADC resolution for small-magnitude inner products,
thereby reducing numerical errors.

The key observation is that
the proposed approach can achieve lower computation errors 
than the original approach.
This improvement results from decomposing each inner product into representative and residual components.
In our approach,
only the residual component is processed by the photonic hardware and 
is therefore affected by DAC/ADC quantization errors and analog photonic noise,
whereas the representative component is computed digitally with high precision.
Since the residual vectors have smaller norms than the corresponding original vectors, 
the error introduced by photonic computation has only a limited effect on the final inner product, 
leading to lower computation errors.

\section{Conclusion}

This paper proposed a matrix-multiplication approach that improves 
the inference accuracy of photonic neural network accelerators by 
exploiting the anisotropy of DNN activation vectors.
The proposed approach reduces the impact of DAC/ADC quantization
errors and analog photonic noise 
without modifying the underlying photonic hardware.
Experimental results on diverse DNN models demonstrated that 
the proposed approach can improve inference accuracy across a
wide range of hardware parameter settings.
These experimental results suggest that exploiting representation anisotropy is a 
promising direction for improving the reliability of 
existing and future photonic neural network accelerators
without requiring hardware modifications.
Future work includes implementing the proposed approach on
photonic accelerator hardware and evaluating its effectiveness
in practical systems.

%\newpage
\section*{Acknowledgments}
This paper is based on results obtained from JST CREST Grant Number JPMJCR24R4.

\bibliographystyle{ACM-Reference-Format}
\bibliography{myrefs}

\end{document}